\documentclass[accepted,xcolor={svgnames}]{uai2022} 

\usepackage[american]{babel}
\usepackage{natbib} 
\usepackage{booktabs} 
\usepackage{tikz} 

\usepackage{url}

\usepackage{graphicx}
\usepackage{amsmath}
\usepackage{amsthm}
\usepackage{booktabs}
\usepackage{hyperref}
\usepackage{algorithm}
\usepackage{algorithmic}
\usepackage{amssymb} 
\usepackage{natbib}
\usepackage{multirow}
\usepackage{diagbox}
\usepackage{nccmath}
\usepackage{ascmac}
\usepackage{here}

\newtheorem{definition}{Definition}
\newtheorem{theorem}{Theorem}

\newtheorem{assumption}{Assumption}
\newtheorem{proposition}{Proposition}
\newtheorem{example}{Example}

\newcommand \indep{\mathop{\perp\!\!\!\!\perp}}

\DeclareMathOperator{\E}{\mathbb{E}}
\DeclareMathOperator{\V}{\mathrm{Var}}
\DeclareMathOperator{\Cov}{\mathrm{Cov}}
\DeclareMathOperator{\R}{\mathbb{R}}
\DeclareMathOperator{\pr}{\mathrm{P}}
\DeclareMathOperator{\I}{\mathbf{I}}

\newcommand{\si}{m}
\DeclareMathOperator{\hzsi}{\mathcal{H}_{0, \si}}
\DeclareMathOperator{\hosi}{\mathcal{H}_{1, \si}}

\usepackage{bm}

\newcommand{\biz}{\textbf{\textit{z}}}

\newcommand{\biX}{\textbf{\textit{X}}}
\newcommand{\bix}{\textbf{\textit{x}}}
\newcommand{\Yz}{Y^{0}}
\newcommand{\Yo}{Y^{1}}
\newcommand{\Ya}{Y^{a}}
\newcommand{\yz}{Y^{0}}
\newcommand{\yzp}{{Y^0}^{\prime}}
\newcommand{\yo}{Y^{1}}
\newcommand{\yop}{{Y^1}^{\prime}}
\newcommand{\wpz}{w^{0}}
\newcommand{\wpo}{w^{1}}
\newcommand{\wpa}{w^{a}}
\newcommand{\xn}{x}
\newcommand{\ozx}{{\omega}^{0, \xn}}
\newcommand{\oox}{{\omega}^{1, \xn}}
\newcommand{\oax}{{\omega}^{a, \xn}}

\newcommand{\xsid}{x_{\si}^{\star}}
\newcommand{\nf}{d}
\newcommand{\nrff}{r}

\newcommand{\acontra}{\Cref{asec:contra}}
\newcommand{\acounter}{\Cref{asec:counter}}
\newcommand{\akernel}{\Cref{asec:kernel}}
\newcommand{\aprop}{\Cref{asubsec:prop1}}
\newcommand{\ath}{\Cref{asubsec:thm1}}
\newcommand{\acounterexp}{\Cref{asubsec:counterexp}}
\newcommand{\aneuronexp}{\Cref{asubsec:neuronexp}}

\usepackage{txfonts}

\usepackage{cleveref}
\crefname{assumption}{assumption}{assumptions}
\crefname{proposition}{proposition}{proposition}
\crefname{example}{example}{example}

\makeatletter
\DeclareRobustCommand\widecheck[1]{{\mathpalette\@widecheck{#1}}}
\def\@widecheck#1#2{%
   \setbox\z@\hbox{\m@th$#1#2$}%
   \setbox\tw@\hbox{\m@th$#1%
      \widehat{%
         \vrule\@width\z@\@height\ht\z@
         \vrule\@height\z@\@width\wd\z@}$}%
   \dp\tw@-\ht\z@
   \@tempdima\ht\z@ \advance\@tempdima2\ht\tw@ \divide\@tempdima\thr@@
   \setbox\tw@\hbox{%
      \raise\@tempdima\hbox{\scalebox{1}[-1]{\lower\@tempdima\box\tw@}}}%
   {\ooalign{\box\tw@ \cr \box\z@}}}
\makeatother

\hypersetup{
    colorlinks=true,
    allcolors=Navy 
}

\title{Feature Selection for Discovering Distributional Treatment Effect Modifiers}

\author[1,2]{Yoichi Chikahara}
\author[2]{Makoto Yamada}
\author[2]{Hisashi Kashima}
\affil[1]{%
    NTT Communication Science Laboratories, Kyoto, Japan
}
\affil[2]{%
    Kyoto University, Kyoto, Japan
}
  
\begin{document}
\maketitle

\begin{abstract}
Finding the features relevant to the difference in treatment effects is essential to unveil the underlying causal mechanisms. Existing methods seek such features by measuring how greatly the feature attributes affect the degree of the {\it conditional average treatment effect} (CATE). However, these methods may overlook important features because CATE, a measure of the average treatment effect, cannot detect differences in distribution parameters other than the mean (e.g., variance). To resolve this weakness of existing methods, we propose a feature selection framework for discovering {\it distributional treatment effect modifiers}. We first formulate a feature importance measure that quantifies how strongly the feature attributes influence the discrepancy between potential outcome distributions. Then we derive its computationally efficient estimator and develop a feature selection algorithm that can control the type I error rate to the desired level. Experimental results show that our framework successfully discovers important features and outperforms the existing mean-based method.
\end{abstract}

\section{Introduction}

When the effects of a treatment (e.g., drug administration) differ across individuals, elucidating why such heterogeneity exists is critical in many applications such as precision medicine \citep{lee2018discovering}, personalized education \citep{schochet2014understanding}, and targeted advertising \citep{taddy2016nonparametric}. A popular approach to explaining treatment effect heterogeneity is to identify the features of an individual that are relevant to the degree of a treatment effect. For instance, to unveil the mechanism of COVID-19 vaccines, recent medical studies have sought the features related to the degree of vaccine-acquired immunity \citep{jabal2021impact}. 

To find such features, we need to measure how greatly the attributes of each feature influence the degree of a treatment effect. To this end, the existing methods use the {\it conditional average treatment effect} (CATE) that is conditioned on each feature, i.e., an average treatment effect across the individuals who have an identical attribute of each feature \citep{imai2013estimating,tian2014simple,zhao2017selective}. However, this average cannot capture distribution parameters other than the mean, such as the variance. As a result, if the attributes of a feature do not affect the average treatment effect but influence other distribution parameters, these mean-based methods will incorrectly conclude that the feature is unrelated to the treatment effect heterogeneity.

The goal of this paper is to propose a feature selection framework for discovering {\it distributional treatment effect modifiers}. To achieve this goal, we develop a feature importance measure that quantifies how greatly the attributes of each feature influence the discrepancy between the distributions of {\it potential outcomes}, i.e., the outcomes when an individual is treated and when not treated. We formulate this measure as a variance of the maximum mean discrepancy (MMD) \citep{gretton2012kernel} between the conditional potential outcome distributions conditioned on each feature. We derive its computationally efficient estimator using a kernel approximation technique and establish a feature selection algorithm that can control the type I error rate (i.e., the proportion of false-positive results) to the desired level. 

{\bf Our contributions} are summarized as follows:
\begin{itemize}
    \item We formulate an MMD-based feature importance measure for discovering distributional treatment effect modifiers (\Cref{UAI2022-sec3-2}). We derive its computationally efficient weighted estimator using a kernel approximation technique (\Cref{UAI2022-sec3-3}). 
    \item We develop an algorithm that selects distributional treatment effect modifiers while controlling the type I error rate (\Cref{UAI2022-sec3-4}). To evaluate significance, we perform multiple hypothesis tests based on the $p$-values computed with the conditional resampling scheme.
    \item We experimentally show that our method successfully finds the features related to treatment effect heterogeneity and outperforms the existing mean-based method.
\end{itemize}

\section{Preliminaries}

\subsection{Problem Setup} \label{UAI2022-sec2-1}

Suppose that we have a sample of $n$ individuals $\mathcal{D} = \{(a_i, \bix_i, y_i)\}_{i=1}^n \overset{i.i.d.}{\sim} \pr(A, \biX, Y)$ for $i = 1, \dots, n$. Here $A \in \{0, 1\}$ is a binary treatment ($A=1$ if an individual is treated; otherwise, $A=0$), $\biX = [X_1, \dots, X_{\nf}]^{\top}$ is $\nf$-dimensional features (a.k.a. covariates), where each feature $X_{\si} \in \mathcal{X}$ ($\si = 1, \dots, \nf$) takes either discrete or continuous values, and $Y \in \R$ is a continuous-valued outcome.\footnote{We assume $Y \in \R$ to use the kernel approximation technique \citep{rahimi2007random}, which is described in \Cref{UAI2022-sec3-3}.} Here we assume that (1) features $\biX$ are measured before applying the treatment and observing outcome $Y$ (i.e., features $\biX$ are \textit{pretreatment variables} and not \textit{mediators} or \textit{colliders} \citep{elwert2014endogenous}) and that (2) features $\biX$ contain all \textit{confounders}, i.e., the variables that affect treatment $A$ and outcome $Y$. Note that these assumptions are standard in the existing work \citep{imai2013estimating,zhao2017selective}.

Given sample $\mathcal{D}$, we solve the problem of selecting the features in $\biX$ that influence the effect of treatment $A$ on outcome $Y$. In this problem, which features should be selected depends on the measurement scale of the treatment effect \citep[Chapter 4]{hernan2020causal}. There are two measurement scales: additive scale $\Yo - \Yz$ and multiplicative scale $\Yo / \Yz$, where $\Yz$ and $\Yo$ are random variables that are referred to as potential outcomes, each of which represents the outcome when $A=0$ and when $A=1$, respectively \citep{rubin1974estimating}. In this study, we define the treatment effect for each individual on an additive scale as $\Yo - \Yz$ because this scale is standard and widely used in numerous applications \citep{lee2018discovering,schochet2014understanding,taddy2016nonparametric}. 

Unfortunately, we cannot observe treatment effect $\Yo - \Yz$. This is because we cannot jointly observe two potential outcomes $\Yz$ and $\Yo$; we only observe either $\Yz$ or $\Yo$, which is obtained as $Y = (1-A) \Yz + A \Yo$ ($A \in \{0, 1\}$). For this reason, existing methods use the average treatment effect across individuals, which can be estimated from the data.

\subsection{Mean-based Approaches} \label{UAI2022-sec2-2}

Many existing methods \citep{tian2014simple,zhao2017selective} seek the features whose attributes affect the degree of the average treatment effect called CATE, which is defined for each feature's attribute, $X_{\si} = \xn$ ($\si = 1, \dots, \nf$), as follows:
\begin{align}
    T_{\si} (\xn) &\coloneqq \E[\Yo - \Yz \mid X_{\si} = \xn] \nonumber \\
    &= \E[\Yo \mid X_{\si} = \xn] - \E[\Yz \mid X_{\si} = \xn]. \label{UAI2022-eq-CATE}
\end{align}
CATE $T_{\si}(\xn)$ is an average treatment effect over the individuals who share an identical attribute, $X_{\si} = \xn$. Note that this CATE is different from the one conditioned on all features $\biX$, which is an inference target of the recent causal inference methods \citep{chang2017informative,hassanpour2019counterfactual,hill2011bayesian,kunzel2019metalearners,nie2021quasi,shalit2017estimating,yoon2018ganite}.

Using CATE $T_{\si}$ ($\si =1, \dots, \nf$), the features that influence the degree of the average treatment effect are defined as the following {\it treatment effect modifiers}:
\begin{definition}[\citet{rothman2008modern}] \label{UAI2022-def_em}
 Feature $X_{\si}$ is said to be a treatment effect modifier if there are at least two values of $X_{\si}$, $x_{\si}$ and $\xsid$ ($x_{\si} \neq \xsid$), such that CATE $T_{\si}$ in \eqref{UAI2022-eq-CATE} takes different values, i.e., $T_{\si}(x_{\si}) \neq T_{\si}(\xsid)$.
\end{definition}
\Cref{UAI2022-def_em} states that feature $X_{\si}$ is a treatment effect modifier if CATE $T_{\si}(\xn)$ is not a constant with respect to value $X_{\si} = \xn$. Roughly speaking, when we group individuals by their $X_{\si}$'s values and compute the average treatment effect in each group of the individuals, if there are at least two groups with different averages, then feature $X_{\si}$ is a treatment effect modifier \citep{vanderweele2009distinction}. 

The existing methods seek such treatment effect modifiers by fitting a regression model that is linear in treatment $A$ with a sparse regularizer \citep{imai2013estimating,sechidis2021using,tian2014simple,zhao2017selective}.

\subsection{Weakness of Mean-based Approaches} \label{UAI2022-sec2-3}

\begin{table}[t]
    \caption{Joint probability tables of potential outcomes in \Cref{UAI2022-ex_1}. Nonzero probabilities are shown in bold. Total expresses marginal potential outcome probabilities.}
    \centering 
        \scalebox{0.785}{
        \begin{tabular}{c|ccc|c}
            \toprule 
            \multicolumn{5}{c}{$\pr(\Yz, \Yo \mid X = 0)$} \\ \midrule
        \diagbox{$\Yz$}{$\Yo$}& -1        & 0    & 1        & Total \\ \midrule 
                            -1&        0  & 0    & 0        & 0 \\
                             0& {\bf 0.5} & 0    & {\bf 0.5}& {\bf 1.0} \\
                             1&        0  & 0    & 0        & 0 \\ \midrule
                         Total& {\bf 0.5} & 0    & {\bf 0.5}& {\bf 1.0} \\
            \bottomrule
        \end{tabular}
        } 
        \scalebox{0.785}{
        \begin{tabular}{c|ccc|c}
            \toprule 
            \multicolumn{5}{c}{$\pr(\Yz, \Yo \mid X = 1)$} \\ \midrule
        \diagbox{$\Yz$}{$\Yo$}& -1 & 0        & 1    & Total \\ \midrule 
                            -1& 0  & 0        & 0    & 0 \\
                             0& 0  & {\bf 1.0}& 0    & {\bf 1.0} \\
                             1& 0  & 0        & 0    & 0 \\ \midrule
                    Total     & 0  & {\bf 1.0}& 0    & {\bf 1.0} \\
            \bottomrule
        \end{tabular}
        }
\label{UAI2022-table_ex_1}

\end{table}

Since the above mean-based methods rely on the average treatment effect, they cannot detect the features whose attributes do not influence the average treatment effect but do affect other functionals of the joint distribution of potential outcomes, such as the covariance between potential outcomes and the treatment effect variance \citep{russell2021sharp}. To illustrate such a feature, consider the following toy example:
\begin{example} \label{UAI2022-ex_1}
    Let $\Yz, \Yo \in \{-1, 0, 1\} \subset \R$ be the potential outcomes and let $X \in \{0, 1\}$ be a binary feature. Suppose that joint distribution $\pr(\Yz, \Yo \mid X)$ is given as \Cref{UAI2022-table_ex_1}. Then feature $X$'s values are irrelevant to the average treatment effect and the covariance between potential outcomes but relevant to the treatment effect variance:
    \begin{align*}
        &\E[\Yo - \Yz \mid X=0] = \E[\Yo - \Yz \mid X=1] = 0\\
        &\Cov[\Yz, \Yo \mid X=0] = \Cov[\Yz, \Yo \mid X=1] = 0\\
        &\V[\Yo - \Yz \mid X=0] = 1; \quad \V[\Yo - \Yz \mid X=1] = 0.
    \end{align*}
\end{example}

Joint distribution $\pr(\Yz, \Yo \mid X)$ presented in \Cref{UAI2022-table_ex_1} shows that feature $X$ is related to a difference in treatment effects: While no individual with attribute $X=1$ receives any treatment effect, those with $X=0$ get positive or negative effects. However, since the CATE values do not depend on $X$, the existing mean-based methods will incorrectly conclude that feature $X$ is unrelated to the treatment effect heterogeneity. This implies that using CATE is insufficient to capture such {\it distributional} treatment effect heterogeneity and might lead to overlooking important features.

\section{Proposed Method}

\subsection{Detecting Distributional Heterogeneity} \label{UAI2022-sec3-1}

We propose a feature selection framework for discovering the features related to distributional treatment effect heterogeneity. To find such features, we consider the problem of determining whether the values of each feature $X_{\si}$ ($\si = 1, \dots, \nf$) influence the functionals of the joint distribution of potential outcomes $\pr(\Yz, \Yo \mid X_{\si})$, such as the average treatment effect, the treatment effect variance, and the covariance between potential outcomes.
\footnote{Identifying which functionals are affected by each feature's values is extremely challenging due to the impossibility of inferring the joint distribution. One possible solution is to use techniques for estimating the lower and upper bounds on these functionals \citep{chen2016inference,russell2021sharp,shingaki2021identification}. Although such bounds require several additional assumptions, they have been successfully applied in several fields, including fairness-aware machine learning \citep{chikahara2021learning}.} This problem is challenging because we cannot infer joint distribution $\pr(\Yz, \Yo \mid X_{\si})$, since we can never jointly observe potential outcomes $\Yz$ and $\Yo$ as described in \Cref{UAI2022-sec2-1}.

To overcome this challenge, we propose measuring the importance of each feature $X_{\si}$ ($\si = 1, \dots, \nf$) by quantifying how greatly $X_{\si}$'s values influence the discrepancy between conditional distributions $\pr(\Yz \mid X_{\si})$ and $\pr(\Yo \mid X_{\si})$. This idea is motivated by the following fact: {\it if the discrepancy between $\pr(\Yz \mid X_{\si})$ and $\pr(\Yo \mid X_{\si})$ varies with $X_{\si}$'s values, then joint distribution $\pr(\Yz, \Yo \mid X_{\si})$ is also changeable depending on $X_{\si}$'s values, and some functionals of the joint distribution depend on $X_{\si}$}. This fact can be easily proved by taking its contraposition, as shown in \acontra.

Such an idea enables us to detect feature $X$ in \Cref{UAI2022-ex_1}, whose values influence the treatment effect variance. This is because, in this example, the discrepancy between conditional potential outcome distributions $\pr(\Yz \mid X)$ and $\pr(\Yo \mid X)$ changes depending on $X$'s values. 

Note, however, that our idea does not always work well. This is because there are counterexamples where feature $X_{\si}$'s values do not affect the discrepancy between conditional distributions $\pr(\Yz \mid X_{\si})$ and $\pr(\Yo \mid X_{\si})$ but influence joint distribution $\pr(\Yz, \Yo \mid X_{\si})$. We take a counterexample in \acounter\ and present the empirical performances in such cases in \acounterexp. Nevertheless, compared with the existing methods, we can detect a wider variety of features relevant to treatment effect heterogeneity, which leads to a better understanding of the underlying causal mechanisms.

\subsection{Feature Importance Measure} \label{UAI2022-sec3-2}

To express the importance of each feature $X_{\si}$ ($\si = 1, \dots, \nf$), we measure the discrepancy between distributions $\pr(\Yz \mid X_{\si})$ and $\pr(\Yo \mid X_{\si})$ using the MMD \citep{gretton2012kernel}.

In fact, there are several MMD-based metrics for measuring the discrepancy between potential outcome distributions \citep{bellot2021kernel,muandet2021counterfactual,park2021conditional}. However, these metrics cannot be applied in our setting because they are not designed for the conditional distributions conditioned on a single feature; we give details of this reason in \Cref{UAI2022-sec5}.

Consequently, we develop an MMD-based metric for conditional distributions $\pr(\Yz \mid X_{\si})$ and $\pr(\Yo \mid X_{\si})$. Let $k_Y\colon \R \times \R \rightarrow \R$ be a positive-definite kernel function. Then the squared MMD between the conditional distributions conditioned on feature value $X_{\si} = \xn$ is defined as
\begin{align}
    &D^2_{\si}(\xn) \coloneqq \mathrm{MMD}^2( \pr(\Yz \mid X_{\si} = \xn), \pr(\Yo \mid X_{\si} = \xn)) \nonumber \\
    = &\E_{\yz, \yzp \mid X_{\si} = X'_{\si} = \xn}[k_Y(\yz, \yzp)] + \E_{\yo, \yop \mid X_{\si} = X'_{\si} = \xn}[k_Y(\yo, \yop)]\nonumber \\
    &- 2 \E_{\yz, \yo \mid X_{\si} = \xn}[k_Y(\yz, \yo)], \label{UAI2022-eq-CDTE} 
\end{align}
where superscript prime $'$ denotes an independent copy of each random variable, and expectation $\E_{\yz, \yzp \mid X_{\si} = X'_{\si} = \xn}$ is taken with respect to $\pr(\Yz, \yzp \mid X_{\si} = X'_{\si} = \xn)$; other expectations are taken in a similar manner. This metric has the following property: If $k_Y$ belongs to the class of kernel functions called {\it characteristic kernels} \citep{gretton2012kernel}, then squared MMD is $D^2_{\si}(\xn) = 0$ if and only if $\pr(\Yz \mid X_{\si} = \xn) = \pr(\Yo \mid X_{\si} = \xn)$. Examples of characteristic kernels include the Gaussian kernel; we provide a brief overview on  characteristic kernels in \akernel.

Based on squared MMD $D^2_{\si}$, we define the features related to distributional treatment effect heterogeneity as the following {\it distributional treatment effect modifiers}:
\begin{definition} \label{UAI2022-def_dem}
    Feature $X_{\si}$ is said to be a distributional treatment effect modifier if there are at least two values of $X_{\si}$, $x_{\si}$ and $\xsid$ ($x_{\si} \neq \xsid$), such that squared MMD $D^2_{\si}$ in \eqref{UAI2022-eq-CDTE} takes different values, i.e., $D^2_{\si}(x_{\si}) \neq D^2_{\si}(\xsid)$.
 \end{definition}
 In other words, feature $X_{\si}$ is a distributional treatment effect modifier if the squared MMD between $\pr(\Yz \mid X_{\si})$ and $\pr(\Yo \mid X_{\si})$ varies depending on $X_{\si}$'s values. 

To detect such a variation, we formulate the importance of each feature $X_{\si}$ as the variance of the squared MMD:
\begin{align}
    I_{\si} &\coloneqq \V[ D^2_{\si}(X_{\si}) ]. \label{UAI2022-eq-imp} 
\end{align}

\subsection{Estimator of Feature Importance} \label{UAI2022-sec3-3}

To estimate feature importance measure $I_{\si}$ in \eqref{UAI2022-eq-imp}, we need to compute the expected values in \eqref{UAI2022-eq-CDTE} whose expectations can be represented as those over conditional distributions $\pr(\Yz \mid X_{\si} = \xn)$ and $\pr(\Yo \mid X_{\si} = \xn)$. 

However, we cannot directly compute them because we have no access to the observations from these conditional distributions. To overcome this difficulty, we develop a weighted estimator that can be computed from the observed data.

\subsubsection{Weighted Conditional MMD (WCMMD)} \label{UAI2022-sec3-3-1}

To infer squared MMD $D^2_{\si}(\xn)$ in \eqref{UAI2022-eq-CDTE}, we develop an estimator of the expected value over conditional distribution $\pr(\Ya \mid X_{\si} = \xn)$ ($a \in \{0, 1\}$) using a weighting-based estimation technique called importance sampling. 

To derive such an estimator, we use weight functions called inverse probability weights \citep{rosenbaum1983central}:
\begin{align}\label{UAI2022-eq-wp}
    \wpz(A, \biX) = \frac{\I(A=0)}{1 - \mathrm{e}(\biX)}, \quad \wpo(A, \biX) = \frac{\I(A=1)}{\mathrm{e}(\biX)},
\end{align} 
where $\mathrm{e}(\biX) \coloneqq \pr(A=1 \mid \biX)$ is the conditional distribution called a {\it propensity score}, and $\I(A=a)$ is an indicator function that takes $1$ if $A = a$; otherwise $0$. In addition, we make the two standard assumptions: {\it positivity}, which imposes support condition $0 < \mathrm{e}(\bix) < 1$ for all $\bix$ \citep{rosenbaum1983central}, and {\it conditional ignorability} (a.k.a. {\it strong ignorability}), which requires conditional independence relation $\{\Yz, \Yo\} \indep A \mid \biX$; this relation is satisfied if features $\biX$ are pretreatment variables, contain no mediator or collider, and include all confounders \citep{elwert2014endogenous}.

Under these assumptions, for instance, expected value $\E_{\Yo \mid X_{\si} = \xn}[\Yo]$ can be reformulated as
\begin{align*}
    &\E_{\Yo \mid X_{\si} = \xn}[\Yo] \\
    = &\E_{\biX_{- \si} \mid X_{\si} = \xn}[\E_{\Yo \mid \biX_{-\si}, X_{\si}=\xn}[\Yo] ] \\
    = &\E_{\biX_{- \si} \mid X_{\si} = \xn, A=1}\left[ \E_{Y \mid \biX_{-\si}, X_{\si}=\xn, A=1}\left[ \frac{\pr(A=1)}{\pr(A=1 \mid \biX)} Y \right] \right] \\
    = &\E_{A, \biX_{- \si}, Y \mid X_{\si} = \xn}[\wpo(A, \biX) Y],
\end{align*}
where $\biX_{-\si} \coloneqq \biX \backslash X_{\si}$ denotes the features with $X_{\si}$ removed.

To estimate squared MMD $D^2_{\si}(\xn)$ in \eqref{UAI2022-eq-CDTE} in the same way, we formulate the following estimator, which we call a {\it weighted conditional MMD} (WCMMD):
\begin{align} 
    &\mbox{WCMMD}^2_{X_{\si} = \xn} \nonumber \\
    \coloneqq &\E_{A, A', \biX_{-\si}, \biX'_{-\si}, Y, Y' \mid X_{\si} = X'_{\si} = \xn}[ \wpz(A, \biX) \wpz(A', \biX') k_Y(Y, Y')] \nonumber \\
    + &\E_{A, A', \biX_{-\si}, \biX'_{-\si}, Y, Y' \mid X_{\si} = X'_{\si} = \xn}[\wpo(A, \biX) \wpo(A', \biX') k_Y(Y, Y')] \nonumber \\
    - &2 \E_{A, A', \biX_{-\si}, \biX'_{-\si}, Y, Y' \mid X_{\si} = X'_{\si} = \xn}[\wpz(A, \biX) \wpo(A', \biX') k_Y(Y, Y')]. \label{UAI2022-eq-Dj}
\end{align}
We can show that this WCMMD equals $D^2_{\si}(\xn)$ under conditional ignorability and positivity assumptions:
\begin{proposition} \label{UAI2022-thm_prop1}
    Suppose that conditional ignorability and positivity hold. Then $D^2_{\si}(\xn) = \mbox{WCMMD}^2_{X_{\si} = \xn}$. 
\end{proposition}
See \aprop\ for the proof. Hence, WCMMD has the same property with $D^2_{\si}(\xn)$: If $k_Y$ is a characteristic kernel, $\mbox{WCMMD}^2_{X_{\si} = \xn} = 0$ if and only if $\pr(\Yz \mid \xn) = \pr(\Yo \mid \xn)$.

\subsubsection{Empirical Estimator of WCMMD}

To infer squared MMD $D^2_{\si}(\xn)$ with estimator \eqref{UAI2022-eq-Dj}, we estimate the conditional expected values conditioned on $X_{\si} = \xn$ using sample $\mathcal{D} = \{(a_i, \bix_i, y_i)\}_{i=1}^n \overset{i.i.d.}{\sim} \pr(A, \biX, Y)$. 

If feature $X_{\si}$ takes discrete values, we only have to take the averages over the individuals with $X_{\si} = \xn$. Formally, by letting $\oax_i$ for $i = 1, \dots, n$ and $a \in \{0, 1\}$ be
\begin{align}
    \oax_i = \frac{\I(x_{\si, i} = \xn)}{\sum_{l=1}^n \I(x_{\si, l} = \xn)} \wpa(a_i, \bix_i) , \label{UAI2022-eq-oax_disc}    
\end{align}
we can estimate the expected values in \eqref{UAI2022-eq-Dj} by
\begin{align}
    \begin{aligned}
        \widehat{D}^2_{\si}(\xn) &\coloneqq \sum_{i=1}^n \sum_{j=1}^n \left(\ozx_i \ozx_j + \oox_i \oox_j \right) k_Y(y_i, y_j) \\
        &- 2 \sum_{i=1}^n \sum_{j=1}^n \ozx_i \oox_j k_Y(y_i, y_j).
    \end{aligned} \label{UAI2022-eq-Dj_}
\end{align}

For continuous-valued feature $X_{\si}$, we smoothen indicator function $\I$ in \eqref{UAI2022-eq-oax_disc} by employing the kernel smoothing technique \citep{nadaraya1964estimating,watson1964smooth} as follows:
\begin{align}
    \oax_i = \frac{ \frac{1}{h_{X_{\si}}} k_{X_{\si}} (x_{\si, i}, \xn)}{\sum_{l=1}^n \frac{1}{h_{X_{\si}}} k_{X_{\si}} (x_{\si, l} , \xn)} \wpa(a_i, \bix_i), \label{UAI2022-eq-oax_cont}
\end{align}
where the similarity between $X_{\si}$'s values is measured by kernel function $k_{X_{\si}}$ with bandwidth $h_{X_{\si}}$; in our experiments, we formulate $k_{X_{\si}}$ as the Gaussian kernel:
\begin{align*}
    k_{X_{\si}}(x_{\si}, x^{\star}_{\si}) = \mathrm{exp}\left(- \frac{\| x_{\si} - x^{\star}_{\si}\|^2}{h^2_{X_{\si}}}\right).
\end{align*}

In both cases where $\oax_i$ is given as \eqref{UAI2022-eq-oax_disc} and \eqref{UAI2022-eq-oax_cont}, we can show the consistency of estimator $\widehat{D}^2_{\si}(\xn)$, i.e., convergence to the true value in the limit of infinite sample size:
\begin{theorem}
    Suppose that weight $\oax_i$ is given as \eqref{UAI2022-eq-oax_disc} or \eqref{UAI2022-eq-oax_cont}. Then under the assumptions presented in \ath, we have $\widehat{D}^2_{\si}(\xn) \overset{p}{\rightarrow} D^2_{\si}(\xn)$ as $n \rightarrow \infty$. \label{UAI2022-thm_thm1}
\end{theorem} 
See \ath\ for the proof. In practice, we need to estimate $\oax_i$ by inferring propensity score $\mathrm{e}(\biX) \coloneqq \pr(A=1 \mid \biX)$ with a regression model (e.g., neural network).

A drawback of estimator $\widehat{D}^2_{\si}(\xn)$ in \eqref{UAI2022-eq-Dj_} is that it needs computation time $O(n^2)$ for sample size $n$, implying that estimating $D^2_{\si}(\xn)$ for each $\xn = x_{\si, 1}, \dots, x_{\si, n}$ requires $O(n^3)$, which is impractical for large $n$. To resolve this issue, in what follows, we develop a computationally efficient variant of $\widehat{D}^2_{\si}(\xn)$.

\subsubsection{Computationally Efficient Empirical Estimator} \label{UAI2022-sec3-3-2}

To reduce the time of computing estimator $\widehat{D}^2_{\si}(\xn)$ in  \eqref{UAI2022-eq-Dj_}, we employ a kernel approximation technique called random Fourier features (RFFs) \citep{rahimi2007random}. 

With RFFs, we approximate kernel function $k_Y(y_i, y_j)$ in \eqref{UAI2022-eq-Dj_} as an inner product of two feature vectors:
\begin{align}
    k_Y(y_i, y_j) \approx \widetilde{k}_Y(y_i, y_j) = \langle \biz(y_i), \biz(y_j) \rangle_{\R^{\nrff}}, \label{UAI2022-eq-rff}
\end{align}
where $\biz\colon \R \rightarrow \R^{\nrff}$ is a mapping that outputs a vector of the $\nrff$ features, where $\nrff$ is a hyperparameter. These $\nrff$ features are randomly sampled from the Fourier transform of kernel function $k_Y$. We formulate $k_Y$ as a Gaussian kernel with bandwidth $h_{Y}$; in this case, feature mapping $\biz$ is given as $\biz(y) = [\sqrt{2} \cos(\lambda_1 y + \zeta_1), \dots, \sqrt{2} \cos(\lambda_{\nrff} y + \zeta_{\nrff})]^{\top}$, where $\lambda_1, \dots, \lambda_{\nrff}$ are drawn from Gaussian distribution $\mathcal{N}(0, 2 h_{Y})$, and $\zeta_1, \dots, \zeta_{\nrff}$ are sampled from uniform distribution $\mathrm{Unif}(0, 2\pi)$, respectively \citep{rahimi2007random}. 

Based on \eqref{UAI2022-eq-rff}, we approximate estimator $\widehat{D}^2_{\si}(\xn)$ in \eqref{UAI2022-eq-Dj_} as
\begin{align}
    \begin{aligned}
    \widetilde{D}^2_{\si}(\xn) &\coloneqq \langle \widetilde{\mu}_{\Yz \mid x}, \widetilde{\mu}_{\Yz \mid x}\rangle_{\mathbb{R}^{\nrff}} + \langle \widetilde{\mu}_{\Yo \mid x}, \widetilde{\mu}_{\Yo \mid x}\rangle_{\mathbb{R}^{\nrff}} \\
    &- 2 \langle \widetilde{\mu}_{\Yz \mid x}, \widetilde{\mu}_{\Yo \mid x}\rangle_{\mathbb{R}^{\nrff}} 
    \end{aligned} \label{UAI2022-eq-hatDj}
\end{align}
where $\widetilde{\mu}_{\Yz \mid x}$ and $\widetilde{\mu}_{\Yo \mid x}$ are the following weighted averages of the $\nrff$-dimensional random feature vector:
\begin{align*}
    \widetilde{\mu}_{\Yz \mid x} = \sum_{i=1}^n \ozx_i \biz(y_i); \ \widetilde{\mu}_{\Yo \mid x} = \sum_{i=1}^n \oox_i \biz(y_i).   
\end{align*}

Using \eqref{UAI2022-eq-hatDj}, we estimate our feature importance measure as
\begin{align}
    \widetilde{I}_{\si} = \frac{1}{n-1} \sum_{\iota=1}^n \left( \widetilde{D}^2_{\si} (x_{\si, \iota}) -  \frac{1}{n} \sum_{\varsigma=1}^n  \widetilde{D}^2_{\si} (x_{\si, \varsigma}) \right)^2. \label{UAI2022-eq-hatimp}
\end{align}
Computing this estimator requires $O(\nrff n^2)$, which is feasible by setting hyperparameter $\nrff$ to a moderate value.

\subsection{Feature Selection with Conditional Randomization Test (CRT)} \label{UAI2022-sec3-4}

Using estimated measures $\widetilde{I}_{1}, \dots, \widetilde{I}_{\nf}$, we select distributional treatment effect modifiers. To achieve this, we perform multiple hypothesis tests where for each $\si = 1, \dots, \nf$, we consider the following null and alternative hypotheses:
\begin{align}
    \hzsi\colon I_{\si} = 0 \quad \mbox{and}\quad  \hosi\colon I_{\si} > 0.
\end{align}

To decide whether to reject each null hypothesis $\hzsi$, we compute $p$-value $p_{\si}$, i.e., the probability of obtaining test statistic $I_{\si}$ such that $I_{\si} \geq \widetilde{I}_{\si}$ under null hypothesis $\hzsi$. Evaluating this $p$-value requires the distribution of test statistic $I_{\si}$ under $\hzsi$. However, analytically deriving this distribution is extremely difficult because the asymptotic distributions of data-dependent weights $\ozx_i$ and $\oox_i$ in feature importance measure $\widetilde{I}_{\si}$ are unclear. 

For this reason, we approximate the distribution of the test statistic under null hypothesis $\hzsi$, where feature $X_{\si}$ is irrelevant to treatment effect heterogeneity. To this end, we simulate such an irrelevant feature for each $X_{\si}$ without changing joint distribution $\pr(\biX)$ so that the joint distribution of this synthetically generated dummy feature and other observed features $\biX_{-\si} \coloneqq \biX \backslash X_{\si}$ is equal to the original joint distribution, $\pr(\biX)$. To achieve this, following the resampling scheme called {\it conditional randomization test} (CRT) \citep[Section F]{candes2018panning}, we sample new $X_{\si}$'s values from the conditional distribution, $\pr(X_{\si} \mid \biX_{- \si})$, without looking at the values of treatment $A$ and outcome $Y$. 

Our CRT proceeds as illustrated in \Cref{UAI2022-alg_CRT}. We first estimate conditional distribution $\pr(X_{\si} \mid \biX_{- \si})$ by fitting a generative model $\mathcal{L}$ to the data; in our experiments, we employ a widely-used deep generative model called the conditional variational autoencoder (CVAE) \citep{sohn2015learning}. Then, using fitted generative model $\mathcal{L}$, we prepare $B$ datasets, each of which contains different values of the synthetic dummy features drawn from $\mathcal{L}$. In particular, for each $b = 1, \dots, B$, we repeat the two steps: sampling $n$ values of feature $X_{\si}$ as $x^{(b)}_{\si, i} \sim \mathcal{L}(X_{\si} \mid \bix_{- \si, i})$ ($i = 1, \dots, n$) and using these values to compute test statistic $\widetilde{I}^{(b)}_{\si}$. By repeating these steps, we obtain an empirical distribution of the test statistic and compute a $p$-value as
\begin{align}
    \hat{p}_{\si} = \frac{1}{B} \sum_{b=1}^B \I \left(\widetilde{I}^{(b)}_{\si} \geq \widetilde{I}_{\si} \right). \label{UAI2022-eq_pval_m}
\end{align}

\begin{algorithm}[t]
    \caption{Conditional Randomization Test (CRT)} \label{UAI2022-alg_CRT}
    \begin{algorithmic}[1]
    \renewcommand{\algorithmicrequire}{\textbf{Input}: sample $\mathcal{D} = \{(a_i, \bix_i, y_i)\}_{i=1}^n$, estimated statistic $\widetilde{I}_{\si}$}
    \renewcommand{\algorithmicensure}{\textbf{Output}: $p$-value $\hat{p}_{\si}$}
    \REQUIRE 
    \ENSURE  
     \STATE Fit generative model $\mathcal{L}$ to sample $\mathcal{D}$. \label{line_deep}
     \FOR {$b = 1, \dots, B$}
        \FOR {$i = 1, \dots, n$}
            \STATE Draw $x^{(b)}_{\si, i} \sim \mathcal{L}(X_{\si} \mid \bix_{- \si, i})$. \label{line_draw}
            \STATE $\bix^{(b)}_i \leftarrow x^{(b)}_{\si, i} \cup \bix_{- \si, i}$ \label{line_xb}
        \ENDFOR
        \STATE Compute test statistic $\widetilde{I}^{(b)}_{\si}$ using $\{(a_i, \bix^{(b)}_i, y_i)\}_{i=1}^n$.  
     \ENDFOR
     \STATE Compute $p$-value $\hat{p}_{\si}$ by Eq. \eqref{UAI2022-eq_pval_m}.
    \RETURN $\hat{p}_{\si}$ 
    \end{algorithmic} 
    \end{algorithm}
    \begin{algorithm}[t]
        \caption{Proposed feature selection framework} \label{UAI2022-alg}
        \begin{algorithmic}[1]
        \renewcommand{\algorithmicrequire}{\textbf{Input}: sample $\mathcal{D} = \{(a_i, \bix_i, y_i)\}_{i=1}^n$, significance level $\alpha$}
        \renewcommand{\algorithmicensure}{\textbf{Output}: feature index set $\hat{S} \subseteq \{1, \dots, \nf\}$}
        \REQUIRE 
        \ENSURE  
         \FOR {$\si = 1, \dots, \nf$}
         \STATE Compute test statistic $\widetilde{I}_{\si}$ with sample $\mathcal{D}$.
         \STATE Compute $p$-value as $\hat{p}_{\si} \leftarrow$ CRT($\mathcal{D}$, $\widetilde{I}_{\si}$).
         \ENDFOR
         \STATE Adjust $p$-values as $\hat{p}^*_1, \dots, \hat{p}^*_{\nf}$ using a multiple testing procedure. \label{line_correct}
         \STATE Select feature index set as $\hat{S} = \{\si \colon \hat{p}^*_{\si} \leq \alpha\}$.
        \RETURN $\hat{S}$ 
        \end{algorithmic} 
    \end{algorithm}

After computing $p$-values $\hat{p}_{1}, \dots, \hat{p}_{\nf}$, we perform multiple hypothesis tests. Since the chance of obtaining false positives increases with the number of hypotheses tested, we control such false positives by adjusting the $p$-values;  we used Benjamini-Hochber (BH) adjustment procedure \citep{benjamini1995controlling} in our experiments. We summarize our feature selection framework in \Cref{UAI2022-alg}.

One of the advantages of applying CRT is that if the fitted generative model equals the true conditional distribution (i.e., $\mathcal{L}(X_{\si} \mid \biX_{- \si}) = \pr(X_{\si} \mid \biX_{- \si})$ for all $\si = 1. \dots, \nf$), it can precisely control the type I error rate to be at most significance level $\alpha$ \citep[Section F]{candes2018panning}. Although learning such generative models is difficult, we experimentally confirmed that our method successfully controlled the type I error rate to be close to $\alpha$ (\Cref{UAI2022-sec4-2}).

As a disadvantage, performing CRT is computationally expensive: It requires computing the test statistic $B$ times for each feature. Although this computation is embarrassingly parallelizable, it needs $O(B \nf \nrff n^2)$ in total, even with our computationally efficient estimator of the test statistic. Our future work will investigate how to further reduce the computation time; for instance, the CRT's computationally efficient variants (e.g., \citet{liu2021fast}) might be helpful.

\section{Experiments} \label{UAI2022-sec4}

\subsection{Setup} \label{UAI2022-sec4-1}

We compared the performance of our proposed framework with the following two baselines: (1) the existing mean-based method called the selective inference method for effect modification (SI-EM) \citep{zhao2017selective} and (2) a naive variant of our method (Naive), which samples the values of a synthetic dummy feature corresponding to $X_{\si}$ ($\si = 1, \dots, \nf$) not from conditional distribution $\pr(X_{\si} \mid \biX_{- \si})$ but from (empirical) marginal distribution $\pr(X_{\si})$. 

We ran all methods with significance level $\alpha = 0.05$. As regards our method and Naive, we set the number of RFFs to $\nrff = 1000$, selected the values of kernel bandwidths $h_{X_1}, \dots, h_{X_{\nf}}$ and $h_Y$ using a well-known heuristic called median heuristic \citep{scholkopf2002learning}, and inferred propensity score $\mathrm{e}(\biX)$ by fitting a feed-forward neural network that contains two linear layers with $50$ neurons and Rectified Linear Unit (ReLU) activation functions. With our method, we performed a CRT by setting the number of resampled datasets to $B = 100$. Here we formulated generative model $\mathcal{L}(X_{\si} \mid \biX_{-\si})$ for each $\si = 1, \dots, \nf$ as a CVAE whose encoders and decoders are given as the feed-forward neural networks that contain two linear layers with $128$ neurons and ReLU functions. We confirmed that the number of neurons did not greatly affect the performance in \aneuronexp.

\subsection{Synthetic Data Experiments} \label{UAI2022-sec4-2}

\begin{figure*}[t]
    \centering
    \includegraphics[width=0.864\textwidth]{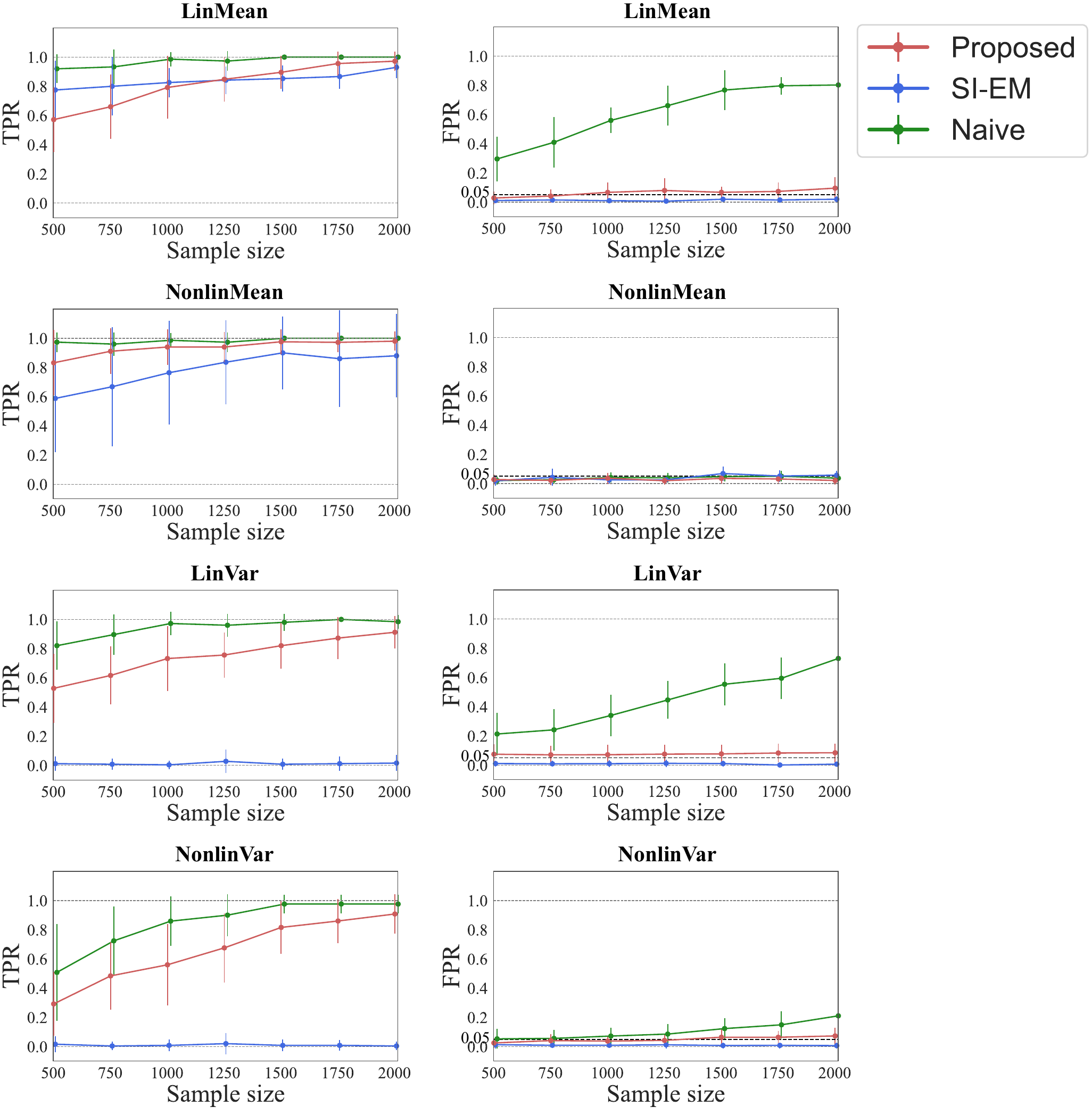}
    \caption{TPRs (left) and FPRs (right) of each method on synthetic data with sample sizes $n = 500, 750, \dots, 2000$. Mean and standard deviation (error bars) over $50$ runs with different datasets are shown.}
    \label{UAI2022-fig_synthTPRFPR}
\end{figure*}

\textbf{Data: } We prepared synthetic datasets as follows. We drew treatment $A$ from the Bernoulli distribution and features $\biX = [X_1, \dots, X_{\nf}]^{\top}$ ($\nf = 30$) from the Gaussian distributions:
\begin{align*}
    &A \sim \mathrm{Ber}(0.5), \\
    &\biX \mid A=0 \sim \mathcal{N}(-\mathbf{\mu}, \mathbf{\Sigma}), \quad \mbox{and} \quad \biX \mid A=1 \sim \mathcal{N}(\mathbf{\mu}, \mathbf{\Sigma}),
\end{align*}
where $\mathrm{Ber}$ and $\mathcal{N}$ denote the Bernoulli and Gaussian distributions, respectively, $\mathbf{\mu} = [0.2, \dots, 0.2]^{\top}$ is a $\nf$-dimensional vector, and $\mathbf{\Sigma} $ is a $\nf \times \nf$ covariance matrix whose $(i, j)$-th element is $\Sigma_{i,j} = \sigma^{| i -j |}$ ($\sigma  = 0.2$) for each $i, j \in \{1, \dots, \nf\}$. We sampled outcome $Y = (1-A) \Yz + A \Yo$ by generating potential outcomes $\Yz$ and $\Yo$ with the following four generation processes where five features $X_1, \dots, X_5$ are distributional treatment effect modifiers:
\begin{itemize}
    \item \textbf{LinMean}: 
    \begin{align*}
        &\Yz \sim \mathcal{N}(-f(X_1, \dots, X_5), 1); \Yo \sim \mathcal{N}( f(X_1, \dots, X_5), 1),
    \end{align*}
    \item \textbf{NonlinMean}: 
    \begin{align*}    
        &\Yz \sim \mathcal{N}(-g(X_1, \dots, X_5), 1); \Yo \sim \mathcal{N}( g(X_1, \dots, X_5), 1),
    \end{align*}
    \item \textbf{LinVar}: 
    \begin{align*}        
    &\Yz \sim \mathcal{N}(-5, 1); \Yo \sim \mathcal{N}( 0, h(f(X_1, \dots, X_5) )^2 ),
    \end{align*}
    \item \textbf{NonlinVar}: 
    \begin{align*}    
    &\Yz \sim \mathcal{N}(-5, 1); \Yo \sim \mathcal{N}( 0, h(g(X_1, \dots, X_5))^2 ),  
    \end{align*}
\end{itemize}
where $f$, $g$ and $h$ are the following functions:
\begin{align*}
    &f(X_1, \dots, X_5) = 4 X_1 + 2 X_2 + X_3 + 2 X_4 + 4 X_5, \\
    &g(X_1, \dots, X_5) = \sum_{j=1}^5 (X_j - 0.5)^3 + 3 \sum_{j=1}^5 X_j - 6, \\
    &h(v) = \mbox{max}(v, 1).
\end{align*}
Under LinMean and NonlinMean, features $X_1, \dots, X_5$ influence the average treatment effect whereas under LinVar and NonlinVar, they affect the treatment effect variance.

\textbf{Results: } Using these synthetic datasets, we evaluated the performance of each method. We computed a true positive rate (TPR) and a false positive rate (FPR), defined as $\frac{\nf_{\mathrm{TP}}}{\nf_{\mathrm{T}}}$ and $\frac{\nf_{\mathrm{FP}}}{\nf - \nf_{\mathrm{T}}}$, where $\nf_{\mathrm{T}} = 5$ is the number of truly relevant features, and $\nf_{\mathrm{TP}}$ and $\nf_{\mathrm{FP}}$ are the number of truly relevant features that are correctly selected as such and the number of irrelevant features that are wrongly selected as the relevant ones, respectively. For each method, we performed $50$ experiments with different synthetic datasets generated with different random numbers and computed the average and the standard deviation of TPRs and FPRs over $50$ runs.

 \Cref{UAI2022-fig_synthTPRFPR} presents the results on the LinMean, NonlinMean, LinVar and NonlinVar datasets. With all of them, our method successfully achieved high TPRs while controlling FPRs to be close to $\alpha = 0.05$. Although SI-EM yielded high TPRs with the LinMean and NonlinMean datasets, since this method is not designed to detect the features related to treatment effect variance, it failed to find important features from the LinVar and NonlinVar datasets. With Naive, not only the TPRs but also the FPRs were higher than our method (especially with the LinMean and LinVar datasets), indicating that it selected many features; however, many of these were false positives, which is problematic in practice. 
 
To further illustrate the difference between our method and Naive, consider how each method approximates the $p$-value of each feature $X_{\si}$ ($\si = 1, \dots, \nf$). Both methods compute the $p$-value by sampling a synthetic dummy feature that is irrelevant to treatment effect heterogeneity; however, its sampling distribution is different. While our method samples it from (estimated) conditional distribution $\pr(X_{\si} \mid \biX_{-\si})$ in the CRT, Naive employs (empirical) marginal distribution $\pr(X_{\si})$ without looking at the values of features $\biX_{-\si}$. The latter generation process \textit{unnecessarily} changes joint distribution $\pr(\biX)$: The joint distribution of a synthetic feature and observed features $\biX_{-\si}$ is greatly different from that of the original features $\biX$; this difference is much larger than with our method. Due to such a large change in $\pr(\biX)$, Naive failed to approximate the test statistic's distribution and yielded high FPRs. By contrast, by avoiding greatly changing joint distribution $\pr(\biX)$ with the CRT, our method effectively evaluated the statistical significance of each feature.
 
Meanwhile, the use of the CRT requires considerable computation time, as discussed in \Cref{UAI2022-sec3-4}. To confirm this, we compared the run time of our method with two baselines: SI-EM and the variant of our method (Exact), which computes the feature importance measure by Eq. \eqref{UAI2022-eq-Dj_} without any approximation. Regarding our method and Exact, we evaluated the total run time, including the training time of the propensity score model and the CVAE. We ran all methods on a 64-bit CentOS machine with 2.10 GHz Xeon Gold 6130 (x2) CPUs and 256-GB RAM. 

\begin{figure}[t]
    \centering
    \includegraphics[width=0.482\textwidth]{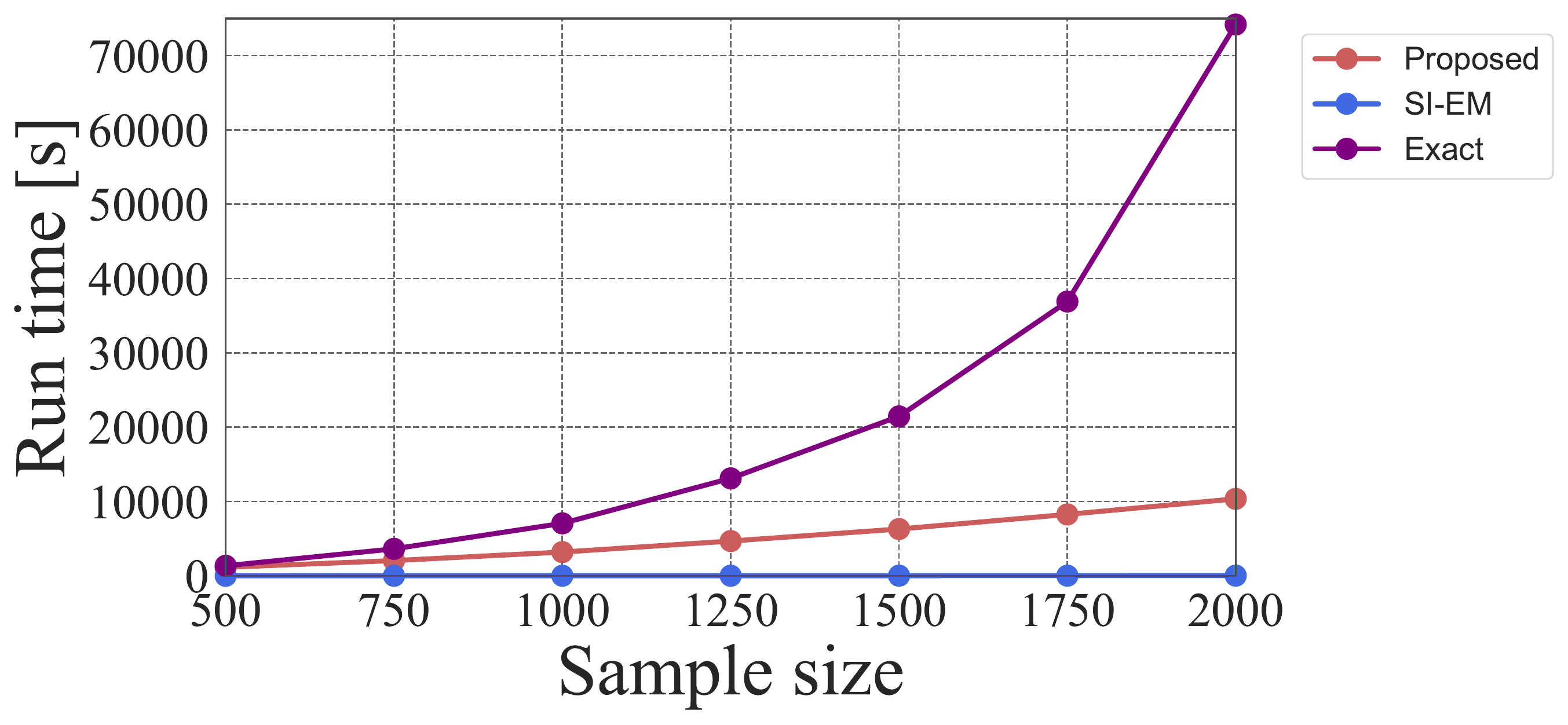}
    \caption{Run time comparison among proposed method (red), SI-EM (blue), and Exact (purple) on LinMean dataset with sample sizes $n = 500, 750, \dots, 2000$}
    \label{UAI2022-fig_runtime}
\end{figure}

\Cref{UAI2022-fig_runtime} shows the run time on the LinMean dataset with sample sizes $n = 500, 750, \dots, 2000$. When $n=2000$, SI-EM and our method required $27$ and $10,360$ seconds, respectively, thus exhibiting a notable difference. However, our method needed far less time than Exact, demonstrating the effectiveness of kernel approximation with RFFs. 

In summary, these results show the following findings:
\begin{itemize}
    \item Our method poses a computational challenge; however, it successfully discovered the features related to the average treatment effect and the treatment effect variance.
    \item SI-EM does not need much time; however, it failed to find the features related to the treatment effect variance.
\end{itemize}
Thus, our proposed feature selection framework has made a significant step toward discovering the features related to distributional treatment effect heterogeneity, which, to the best of our knowledge, is the first attempt in causal inference studies. A further reduction of computation time is left as our future work, as described in \Cref{UAI2022-sec3-4}.

\begin{table}[t]
    \caption{$p$-values of features selected by our method from NHANES dataset: Mean and standard deviation are shown for all features with mean $p$-values less than $\alpha = 0.05$.}

        \begin{tabular}{l|l}
            \toprule 
        Feature & Adjusted $p$-value \\ \midrule 
                            Age &  $0.0075 \pm 0.0305$      \\
                            Gender & $0.0046 \pm 0.0269$       \\
                            Number of cigarettes smoked & $0.0 \pm 0.0$   \\ \bottomrule
        \end{tabular}
\label{UAI2022-table_real}

\end{table}

\subsection{Real-World Data Experiments} \label{UAI2022-sec4-3}

\textbf{Data: } We used the health records from the National Health and Nutrition Examination Survey (NHANES).\footnote{\url{https://wwwn.cdc.gov/nchs/nhanes/}} Following \citet{zhao2017selective}, we collected the records of $n = 9677$ individuals. Each record contains $\nf = 20$ features, such as age, gender, race, income, and past medical history (e.g., asthma, gout, stroke, and heart disease); $3$ of them take continuous values, and the others are discrete.

With this dataset, we investigated which features modify the effects of obesity on low-grade systemic inflammation by regarding whether body mass index (BMI) exceeds $25$ as treatment $A$ and serum C-reactive protein (CRP) level as outcome $Y$. Discovering such features has important medical implications because low-grade inflammation increases the risk of various chronic diseases, such as cancers and cardiovascular disease \citep{rodriguez2013obesity}.

Since the truly relevant features are unknown, we cannot evaluate the TPRs and FPRs. For this reason, we compared the features selected by our method and SI-EM. Since our method is founded on the randomized algorithm (i.e., CRT), we computed the mean of the adjusted $p$-values over $50$ runs and used this mean $p$-value to select the features.

\textbf{Results: } \Cref{UAI2022-table_real} presents the adjusted $p$-values for all features that are selected by our proposed method.

Both our method and SI-EM successfully selected age and gender, which were reported as important in the previous medical studies \citep{visser1999elevated}. Although SI-EM selected only these two features, our method concluded that the number of cigarettes smoked is also statistically significant. Selecting this feature is interesting and seems reasonable because the synergistic effect of obesity and smoking on systemic inflammation has been reported in previous studies \citep{olafsdottir2005c}.

\section{Related Work} \label{UAI2022-sec5}

\textbf{Interpreting treatment effect heterogeneity: } A growing number of causal inference methods have been developed to accurately estimate heterogeneous treatment effects using neural networks \citep{johansson2016learning,shalit2017estimating,yoon2018ganite}, tree-based models \citep{hahn2020bayesian,hill2011bayesian}, and machine learning frameworks called meta-learners \citep{kunzel2019metalearners,nie2021quasi}. 

However, few are designed to elucidate a causal mechanism that yields the treatment effect heterogeneity. The Causal Rule Ensemble method \citep{lee2020causal} seeks the important features by learning a rule-based model that emulates the input-output relationship of a fitted treatment effect estimation model. \citet{gilad2021heterogeneous} considered a hypothesis test for discovering the treatment effect modifiers from social network data. However, none of these methods can find the features related to distributional treatment effect heterogeneity because they are also based on the average treatment effect and cannot find the features related to other functionals of the joint distribution of potential outcomes.

To overcome this limitation of the existing mean-based methods, we established a feature selection framework for discovering the important features related to the functionals of the joint distribution of potential outcomes.

\textbf{MMD between potential outcome distributions: } To find distributional treatment effect modifiers, we formulated a weighted estimator of the MMD that measures the discrepancy between conditional potential outcome distributions.

Our estimator has a clear advantage in that it can consistently estimate the MMD between the conditional distributions conditioned on a single feature, $\pr(\Yz \mid X_{\si})$ and $\pr(\Yo \mid X_{\si})$ ($\si = 1, \dots, \nf$), by addressing the confounders in features $\biX$. 

The existing estimators cannot consistently estimate such an MMD. The kernel treatment effect (KTE) \citep{muandet2021counterfactual} and the weighted MMD (WMMD) \citep{bellot2021kernel} are designed to quantify the discrepancy between marginal distributions $\pr(\Yz)$ and $\pr(\Yo)$; hence they cannot address the conditional distributions. Although the conditional distributional treatment effect (CoDiTE) \citep{park2021conditional} measures the MMD between conditional distributions $\pr(\Yz \mid \biX)$ and $\pr(\Yo \mid \biX)$, we cannot naively apply it by considering the setting where features $\biX$ only contain a single feature (i.e., $\biX = \{X_{\si}\}$). This is because this measure only addresses the confounders that are included in the conditioning variables, and if setting $\biX = \{X_{\si}\}$, we cannot eliminate the influence of the confounders in $\biX_{-\si}$.

To consistently estimate the MMD between conditional distributions $\pr(\Yz \mid X_{\si})$ and $\pr(\Yo \mid X_{\si})$, we derived an IPW-based estimator by regarding the MMD as a function of features $\biX$ and then averaging out unwanted features $\biX_{-\si}$ (by taking an integral with respect to $\pr(\biX_{-\si} \mid X_{\si})$).

\section{Conclusion}

We proposed a feature selection framework for discovering the features related to the distributional treatment effect heterogeneity. The key advantage of our framework is that it can identify the features whose values influence the functionals of the joint distribution of potential outcomes if the feature values also affect the discrepancy between conditional potential outcome distributions. To the best of our knowledge, this is the first feature selection approach to revealing the causal mechanism that yields the distributional treatment effect heterogeneity. We experimentally show that our feature selection framework successfully selected important features and outperformed the existing method.

\bibliography{main}

\begin{thebibliography}{50}
\providecommand{\natexlab}[1]{#1}
\providecommand{\url}[1]{\texttt{#1}}
\expandafter\ifx\csname urlstyle\endcsname\relax
  \providecommand{\doi}[1]{doi: #1}\else
  \providecommand{\doi}{doi: \begingroup \urlstyle{rm}\Url}\fi

\bibitem[Bellot and van~der Schaar(2021)]{bellot2021kernel}
Alexis Bellot and Mihaela van~der Schaar.
\newblock A kernel two-sample test with selection bias.
\newblock In \emph{UAI}, 2021.

\bibitem[Benjamini and Hochberg(1995)]{benjamini1995controlling}
Yoav Benjamini and Yosef Hochberg.
\newblock Controlling the false discovery rate: {A} practical and powerful
  approach to multiple testing.
\newblock \emph{Journal of Royal statistical society: series B
  (Methodological)}, 57\penalty0 (1):\penalty0 289--300, 1995.

\bibitem[Candes et~al.(2018)Candes, Fan, Janson, and Lv]{candes2018panning}
Emmanuel Candes, Yingying Fan, Lucas Janson, and Jinchi Lv.
\newblock {Panning for gold:‘{M}odel-X’knockoffs for high dimensional
  controlled variable selection}.
\newblock \emph{Journal of Royal Statistical Society: Series B (Statistical
  Methodology)}, 80\penalty0 (3):\penalty0 551--577, 2018.

\bibitem[Chang and Dy(2017)]{chang2017informative}
Yale Chang and Jennifer Dy.
\newblock Informative subspace learning for counterfactual inference.
\newblock In \emph{AAAI}, pages 1770--1776, 2017.

\bibitem[Chen et~al.(2016)Chen, Fan, and Liu]{chen2016inference}
Heng Chen, Yanqin Fan, and Ruixuan Liu.
\newblock {Inference for the correlation coefficient between potential outcomes
  in the Gaussian switching regime model}.
\newblock \emph{Journal of Econometrics}, 195\penalty0 (2):\penalty0 255--270,
  2016.

\bibitem[Chikahara et~al.(2021)Chikahara, Sakaue, Fujino, and
  Kashima]{chikahara2021learning}
Yoichi Chikahara, Shinsaku Sakaue, Akinori Fujino, and Hisashi Kashima.
\newblock Learning individually fair classifier with path-specific
  causal-effect constraint.
\newblock In \emph{AISTATS}, pages 145--153, 2021.

\bibitem[Elwert and Winship(2014)]{elwert2014endogenous}
Felix Elwert and Christopher Winship.
\newblock Endogenous selection bias: {T}he problem of conditioning on a
  collider variable.
\newblock \emph{Annual Review of Sociology}, 40:\penalty0 31--53, 2014.

\bibitem[Gilad et~al.(2021)Gilad, Parikh, Roy, and
  Salimi]{gilad2021heterogeneous}
Amir Gilad, Harsh Parikh, Sudeepa Roy, and Babak Salimi.
\newblock Heterogeneous treatment effects in social networks.
\newblock \emph{arXiv preprint arXiv:2105.10591}, 2021.

\bibitem[Gretton et~al.(2012)Gretton, Borgwardt, Rasch, Sch{\"o}lkopf, and
  Smola]{gretton2012kernel}
Arthur Gretton, Karsten~M. Borgwardt, Malte~J Rasch, Bernhard Sch{\"o}lkopf,
  and Alexander Smola.
\newblock A kernel two-sample test.
\newblock \emph{JMLR}, 13\penalty0 (1):\penalty0 723--773, 2012.

\bibitem[Hahn et~al.(2020)Hahn, Murray, and Carvalho]{hahn2020bayesian}
P.~Richard Hahn, Jared~S. Murray, and Carlos~M. Carvalho.
\newblock Bayesian regression tree models for causal inference:
  {R}egularization, confounding, and heterogeneous effects.
\newblock \emph{Bayesian Analysis}, 15\penalty0 (3):\penalty0 965--1056, 2020.

\bibitem[Hassanpour and Greiner(2019)]{hassanpour2019counterfactual}
Negar Hassanpour and Russell Greiner.
\newblock Counterfactual regression with importance sampling weights.
\newblock In \emph{IJCAI}, pages 5880--5887, 2019.

\bibitem[Hern\'{a}n and Robins(2020)]{hernan2020causal}
Miguel~A. Hern\'{a}n and James~M. Robins.
\newblock \emph{Causal Inference: What if. Boca Raton: Chapman \& Hill/CRC}.
\newblock 2020.

\bibitem[Hill(2011)]{hill2011bayesian}
Jennifer~L. Hill.
\newblock Bayesian nonparametric modeling for causal inference.
\newblock \emph{Journal of Computational and Graphical Statistics}, 20\penalty0
  (1):\penalty0 217--240, 2011.

\bibitem[Hoeffding(1961)]{hoeffding1961strong}
Wassily Hoeffding.
\newblock The strong law of large numbers for {U}-statistics.
\newblock Technical report, 1961.

\bibitem[Imai and Ratkovic(2013)]{imai2013estimating}
Kosuke Imai and Marc Ratkovic.
\newblock Estimating treatment effect heterogeneity in randomized program
  evaluation.
\newblock \emph{Annals of Applied Statistics}, 7\penalty0 (1):\penalty0
  443--470, 2013.

\bibitem[Jabal et~al.(2021)Jabal, Ben-Amram, Beiruti, Batheesh, Sussan, Zarka,
  and Edelstein]{jabal2021impact}
Kamal~Abu Jabal, Hila Ben-Amram, Karine Beiruti, Yunis Batheesh, Christian
  Sussan, Salman Zarka, and Michael Edelstein.
\newblock {Impact of age, ethnicity, sex and prior infection status on
  immunogenicity following a single dose of the BNT162b2 mRNA COVID-19 vaccine:
  {R}eal-world evidence from healthcare workers, Israel, December 2020 to
  January 2021}.
\newblock \emph{Eurosurveillance}, 26\penalty0 (6), 2021.

\bibitem[Johansson et~al.(2016)Johansson, Shalit, and
  Sontag]{johansson2016learning}
Fredrik Johansson, Uri Shalit, and David Sontag.
\newblock Learning representations for counterfactual inference.
\newblock In \emph{ICML}, pages 3020--3029, 2016.

\bibitem[K{\"u}nzel et~al.(2019)K{\"u}nzel, Sekhon, Bickel, and
  Yu]{kunzel2019metalearners}
S{\"o}ren~R. K{\"u}nzel, Jasjeet~S. Sekhon, Peter~J. Bickel, and Bin Yu.
\newblock Metalearners for estimating heterogeneous treatment effects using
  machine learning.
\newblock \emph{Proceedings of National Academy of Sciences}, 116\penalty0
  (10):\penalty0 4156--4165, 2019.

\bibitem[Lee et~al.(2018)Lee, Small, Hsu, Silber, and
  Rosenbaum]{lee2018discovering}
Kwonsang Lee, Dylan~S. Small, Jesse~Y. Hsu, Jeffrey~H. Silber, and Paul~R.
  Rosenbaum.
\newblock Discovering effect modification in an observational study of surgical
  mortality at hospitals with superior nursing.
\newblock \emph{Journal of Royal Statistical Society: Series A (Statistics in
  Society)}, 181\penalty0 (2):\penalty0 535--546, 2018.

\bibitem[Lee et~al.(2020)Lee, Bargagli-Stoffi, and Dominici]{lee2020causal}
Kwonsang Lee, Falco~J. Bargagli-Stoffi, and Francesca Dominici.
\newblock Causal rule ensemble: {I}nterpretable inference of heterogeneous
  treatment effects.
\newblock \emph{arXiv preprint arXiv:2009.09036}, 2020.

\bibitem[Liu et~al.(2021)Liu, Katsevich, Janson, and Ramdas]{liu2021fast}
Molei Liu, Eugene Katsevich, Lucas Janson, and Aaditya Ramdas.
\newblock {Fast and powerful conditional randomization testing via
  distillation}.
\newblock \emph{Biometrika}, 2021.

\bibitem[Muandet et~al.(2017)Muandet, Fukumizu, Sriperumbudur, Sch{\"o}lkopf,
  et~al.]{muandet2017kernel}
Krikamol Muandet, Kenji Fukumizu, Bharath Sriperumbudur, Bernhard
  Sch{\"o}lkopf, et~al.
\newblock Kernel mean embedding of distributions: {A} review and beyond.
\newblock \emph{Foundations and Trends{\textregistered} in Machine Learning},
  10\penalty0 (1-2):\penalty0 1--141, 2017.

\bibitem[Muandet et~al.(2021)Muandet, Kanagawa, Saengkyongam, and
  Marukatat]{muandet2021counterfactual}
Krikamol Muandet, Motonobu Kanagawa, Sorawit Saengkyongam, and Sanparith
  Marukatat.
\newblock Counterfactual mean embeddings.
\newblock \emph{JMLR}, 22\penalty0 (162):\penalty0 1--71, 2021.

\bibitem[Nadaraya(1964)]{nadaraya1964estimating}
Elizbar~A. Nadaraya.
\newblock On estimating regression.
\newblock \emph{Theory of Probability \& Its Applications}, 9\penalty0
  (1):\penalty0 141--142, 1964.

\bibitem[Nie and Wager(2021)]{nie2021quasi}
Xinkun Nie and Stefan Wager.
\newblock Quasi-oracle estimation of heterogeneous treatment effects.
\newblock \emph{Biometrika}, 108\penalty0 (2):\penalty0 299--319, 2021.

\bibitem[{\'O}lafsd{\'o}ttir et~al.(2005){\'O}lafsd{\'o}ttir, Gislason,
  Thjodleifsson, Olafsson, Gislason, J{\"o}gi, and Janson]{olafsdottir2005c}
Inga~Sif {\'O}lafsd{\'o}ttir, Thorarinn Gislason, B.~Thjodleifsson,
  I.~Olafsson, D.~Gislason, Rain J{\"o}gi, and Christer Janson.
\newblock C reactive protein levels are increased in non-allergic but not
  allergic asthma: {A} multicentre epidemiological study.
\newblock \emph{Thorax}, 60\penalty0 (6):\penalty0 451--454, 2005.

\bibitem[Park et~al.(2021)Park, Shalit, Sch{\"o}lkopf, and
  Muandet]{park2021conditional}
Junhyung Park, Uri Shalit, Bernhard Sch{\"o}lkopf, and Krikamol Muandet.
\newblock {Conditional distributional treatment effect with kernel conditional
  mean embeddings and U-statistic regression}.
\newblock In \emph{ICML}, pages 8401--8412, 2021.

\bibitem[Rahimi et~al.(2007)Rahimi, Recht, et~al.]{rahimi2007random}
Ali Rahimi, Benjamin Recht, et~al.
\newblock Random features for large-scale kernel machines.
\newblock In \emph{NeurIPS}, volume~3, page~5, 2007.

\bibitem[Rodr{\'\i}guez-Hern{\'a}ndez et~al.(2013)Rodr{\'\i}guez-Hern{\'a}ndez,
  Simental-Mend{\'\i}a, Rodr{\'\i}guez-Ram{\'\i}rez, and
  Reyes-Romero]{rodriguez2013obesity}
Heriberto Rodr{\'\i}guez-Hern{\'a}ndez, Luis~E. Simental-Mend{\'\i}a, Gabriela
  Rodr{\'\i}guez-Ram{\'\i}rez, and Miguel~A. Reyes-Romero.
\newblock Obesity and inflammation: {E}pidemiology, risk factors, and markers
  of inflammation.
\newblock \emph{International journal of endocrinology}, 2013.

\bibitem[Rosenbaum and Rubin(1983)]{rosenbaum1983central}
Paul~R. Rosenbaum and Donald~B. Rubin.
\newblock The central role of the propensity score in observational studies for
  causal effects.
\newblock \emph{Biometrika}, 70\penalty0 (1):\penalty0 41--55, 1983.

\bibitem[Rothman et~al.(2008)Rothman, Greenland, Lash,
  et~al.]{rothman2008modern}
Kenneth~J. Rothman, Sander Greenland, Timothy~L. Lash, et~al.
\newblock \emph{{Modern Epidemiology}}, volume~3.
\newblock Wolters Kluwer Health/Lippincott Williams \& Wilkins Philadelphia,
  2008.

\bibitem[Rubin(1974)]{rubin1974estimating}
Donald~B. Rubin.
\newblock Estimating causal effects of treatments in randomized and
  nonrandomized studies.
\newblock \emph{Journal of Educational Psychology}, 66\penalty0 (5):\penalty0
  688, 1974.

\bibitem[Russell(2021)]{russell2021sharp}
Thomas~M. Russell.
\newblock Sharp bounds on functionals of the joint distribution in the analysis
  of treatment effects.
\newblock \emph{Journal of Business \& Economic Statistics}, 39\penalty0
  (2):\penalty0 532--546, 2021.

\bibitem[Schochet et~al.(2014)Schochet, Puma, and
  Deke]{schochet2014understanding}
Peter~Z. Schochet, Mike Puma, and John Deke.
\newblock Understanding variation in treatment effects in education impact
  evaluations: {A}n overview of quantitative methods.
\newblock \emph{National Center for Education Evaluation and Regional
  Assistance}, 2014.

\bibitem[Sch{\"o}lkopf et~al.(2002)Sch{\"o}lkopf, Smola, Bach,
  et~al.]{scholkopf2002learning}
Bernhard Sch{\"o}lkopf, Alexander~J. Smola, Francis Bach, et~al.
\newblock \emph{Learning with kernels: {S}upport vector machines,
  regularization, optimization, and beyond}.
\newblock MIT press, 2002.

\bibitem[Sechidis et~al.(2021)Sechidis, Kormaksson, and
  Ohlssen]{sechidis2021using}
Konstantinos Sechidis, Matthias Kormaksson, and David Ohlssen.
\newblock Using knockoffs for controlled predictive biomarker identification.
\newblock \emph{Statistics in Medicine}, 40\penalty0 (25):\penalty0 5453--5473,
  2021.

\bibitem[Serfling(2009)]{serfling2009approximation}
Robert~J. Serfling.
\newblock \emph{Approximation theorems of mathematical statistics}, volume 162.
\newblock John Wiley \& Sons, 2009.

\bibitem[Shalit et~al.(2017)Shalit, Johansson, and
  Sontag]{shalit2017estimating}
Uri Shalit, Fredrik~D. Johansson, and David Sontag.
\newblock Estimating individual treatment effect: {G}eneralization bounds and
  algorithms.
\newblock In \emph{ICML}, pages 3076--3085, 2017.

\bibitem[Shingaki and Kuroki(2021)]{shingaki2021identification}
Ryusei Shingaki and Manabu Kuroki.
\newblock Identification and estimation of joint probabilities of potential
  outcomes in observational studies with covariate information.
\newblock In \emph{NeurIPS}, 2021.

\bibitem[Smola et~al.(2007)Smola, Gretton, Song, and
  Sch{\"o}lkopf]{smola2007hilbert}
Alex Smola, Arthur Gretton, Le~Song, and Bernhard Sch{\"o}lkopf.
\newblock A {H}ilbert space embedding for distributions.
\newblock In \emph{International Conference on Algorithmic Learning Theory},
  pages 13--31, 2007.

\bibitem[Sohn et~al.(2015)Sohn, Lee, and Yan]{sohn2015learning}
Kihyuk Sohn, Honglak Lee, and Xinchen Yan.
\newblock Learning structured output representation using deep conditional
  generative models.
\newblock In \emph{NeurIPS}, pages 3483--3491, 2015.

\bibitem[Sriperumbudur et~al.(2010)Sriperumbudur, Gretton, Fukumizu,
  Sch{\"o}lkopf, and Lanckriet]{sriperumbudur2010hilbert}
Bharath~K. Sriperumbudur, Arthur Gretton, Kenji Fukumizu, Bernhard
  Sch{\"o}lkopf, and Gert~R.G. Lanckriet.
\newblock Hilbert space embeddings and metrics on probability measures.
\newblock \emph{JMLR}, 11:\penalty0 1517--1561, 2010.

\bibitem[Taddy et~al.(2016)Taddy, Gardner, Chen, and
  Draper]{taddy2016nonparametric}
Matt Taddy, Matt Gardner, Liyun Chen, and David Draper.
\newblock A nonparametric bayesian analysis of heterogenous treatment effects
  in digital experimentation.
\newblock \emph{Journal of Business \& Economic Statistics}, 34\penalty0
  (4):\penalty0 661--672, 2016.

\bibitem[Tian et~al.(2014)Tian, Alizadeh, Gentles, and
  Tibshirani]{tian2014simple}
Lu~Tian, Ash~A. Alizadeh, Andrew~J. Gentles, and Robert Tibshirani.
\newblock A simple method for estimating interactions between a treatment and a
  large number of covariates.
\newblock \emph{Journal of American Statistical Association}, 109\penalty0
  (508):\penalty0 1517--1532, 2014.

\bibitem[VanderWeele(2009)]{vanderweele2009distinction}
Tyler~J. VanderWeele.
\newblock On the distinction between interaction and effect modification.
\newblock \emph{Epidemiology}, 20\penalty0 (6):\penalty0 863--871, 2009.

\bibitem[Visser et~al.(1999)Visser, Bouter, McQuillan, Wener, and
  Harris]{visser1999elevated}
Marjolein Visser, Lex~M. Bouter, Geraldine~M. McQuillan, Mark~H. Wener, and
  Tamara~B. Harris.
\newblock {Elevated C-reactive protein levels in overweight and obese adults}.
\newblock \emph{Journal of Americal Medical Association}, 282\penalty0
  (22):\penalty0 2131--2135, 1999.

\bibitem[Watson(1964)]{watson1964smooth}
Geoffrey~S. Watson.
\newblock Smooth regression analysis.
\newblock \emph{Sankhy{\=a}: Indian Journal of Statistics, Series A}, pages
  359--372, 1964.

\bibitem[Wied and Wei{\ss}bach(2012)]{wied2012consistency}
Dominik Wied and Rafael Wei{\ss}bach.
\newblock Consistency of the kernel density estimator: {A} survey.
\newblock \emph{Statistical Papers}, 53\penalty0 (1):\penalty0 1--21, 2012.

\bibitem[Yoon et~al.(2018)Yoon, Jordon, and Van Der~Schaar]{yoon2018ganite}
Jinsung Yoon, James Jordon, and Mihaela Van Der~Schaar.
\newblock {GANITE: {E}stimation of individualized treatment effects using
  generative adversarial nets}.
\newblock In \emph{ICLR}, 2018.

\bibitem[Zhao et~al.(2022)Zhao, Small, and Ertefaie]{zhao2017selective}
Qingyuan Zhao, Dylan~S. Small, and Ashkan Ertefaie.
\newblock Selective inference for effect modification via the lasso.
\newblock \emph{Journal of Royal Statistical Society: Series B (Statistical
  Methodology)}, 84\penalty0 (2):\penalty0 382--413, 2022.

\end{thebibliography}

\clearpage
\appendix
\setcounter{theorem}{0}
\setcounter{proposition}{0}

\title{Feature Selection for Discovering Distributional Treatment Effect Modifiers (Supplementary material)}

\onecolumn

\maketitle

\section{Relationship between Marginal and Joint Distributions} \label{asec:contra}

To confirm that our feature importance measure is reasonable, we consider the following two relationships:
\begin{screen}
    \begin{itemize}
        \item {\it If the discrepancy between marginal potential outcome distributions $\pr(\Yz \mid X_{\si})$ and $\pr(\Yo \mid X_{\si})$ varies with feature $X_{\si}$'s values, then joint distribution $\pr(\Yz, \Yo \mid X_{\si})$ is also changeable depending on $X_{\si}$'s values}.
        \item {\it If joint distribution $\pr(\Yz, \Yo \mid X_{\si})$ changes depending on feature $X_{\si}$'s values, then some functionals of the joint distribution depend on $X_{\si}$'s values}.
    \end{itemize}
\end{screen}
Since the second relationship is obvious, in this section, we show that the first relationship holds. For simplicity, we consider binary feature $X_{\si} \in \{0, 1\}$; however, the following discussion also holds for discrete-valued and continuous-valued $X_{\si}$. 

To prove the first relationship, it is sufficient to show that its contraposition holds: If $\pr(\Yz, \Yo \mid X_{\si} = 0) = \pr(\Yz, \Yo \mid X_{\si} = 1)$, then the discrepancy between $\pr(\Yz \mid X_{\si} = 0)$ and $\pr(\Yo \mid X_{\si} = 0)$ equals the one between $\pr(\Yz \mid X_{\si} = 1)$ and $\pr(\Yo \mid X_{\si} = 1)$. We can easily prove this contraposition. From the equality of the joint distributions, we have $\pr(\Yz \mid X_{\si} = 0) = \pr(\Yz \mid X_{\si} = 1)$ and $\pr(\Yo \mid X_{\si} = 0) = \pr(\Yo \mid X_{\si} = 1)$. These equalities imply that the discrepancy between $\pr(\Yz \mid X_{\si} = 0)$ and $\pr(\Yo \mid X_{\si} = 0)$ equals the one between $\pr(\Yz \mid X_{\si} = 1)$ and $\pr(\Yo \mid X_{\si} = 1)$. Thus we proved the first relationship.

\section{Counterexamples} \label{asec:counter}
 
As described in \Cref{UAI2022-sec3-1}, there are several counterexamples where our method cannot find the features related to the functionals of the joint distribution of potential outcomes. 

Let $\Yz$ and $\Yo$ be the potential outcomes and $X \in \{0, 1\}$ be a binary feature. Suppose that the discrepancy between marginal distributions $\pr(\Yz \mid X)$ and $\pr(\Yo \mid X)$ is measured as the MMD \citep{gretton2012kernel}. Then we can represent such counterexamples as the cases where the following relations hold:
\begin{align*}
    &\pr(\Yz, \Yo \mid X=0) \neq  \pr(\Yz, \Yo \mid X=1) \\
    &\mathrm{MMD}^2( \pr(\Yz \mid X = 0), \pr(\Yo \mid X = 0)) = \mathrm{MMD}^2( \pr(\Yz \mid X = 1), \pr(\Yo \mid X = 1)).
\end{align*}

Letting the potential outcomes be $\Yz, \Yo \in \{-1, 0, 1\} \subset \R$, we take an example of joint probability tables that satisfies the above relations in \Cref{UAI2022-table_ex_2}. In this example, the MMD between marginal distributions remains unchanged:
\begin{align*}
    &\mathrm{MMD}^2( \pr(\Yz \mid X = 0), \pr(\Yo \mid X = 0)) = \mathrm{MMD}^2( \pr(\Yz \mid X = 1), \pr(\Yo \mid X = 1)) = 0.
\end{align*}
By contrast, the joint distribution changes depending on $X$'s values, as illustrated in \Cref{UAI2022-table_ex_2}. As a result, although the average treatment effect does not change, the treatment effect variance and the covariance between potential outcomes vary as follows:
\begin{align*}
    &\E[\Yo - \Yz \mid X=0] = \E[\Yo - \Yz \mid X=1] = 0\\
    &\Cov[\Yz, \Yo \mid X=0] = 1; \quad \Cov[\Yz, \Yo \mid X=1] = -1\\
    &\V[\Yo - \Yz \mid X=0] = 0; \quad \V[\Yo - \Yz \mid X=1] = 4.
\end{align*}

In this example, since we cannot detect any change in the MMD between marginal distributions, our method fails to find that feature $X$ is related to treatment effect heterogeneity. Note, however, that the existing mean-based approaches would also fail because the average treatment effect remains unchanged. 

Addressing such counterexamples is extremely difficult. It requires us to estimate the functionals of the joint potential outcome distribution; however, inferring such a joint distribution is impossible, as described in \Cref{UAI2022-sec3-1}. One possible solution is to utilize several techniques for estimating the lower and upper bounds on these functionals by making additional assumptions \citep{chen2016inference,russell2021sharp,shingaki2021identification}. Establishing a feature selection framework that utilizes such lower and upper bounds remains our future work.

\begin{table}[t]
    \caption{Joint probability tables of potential outcomes. Nonzero probabilities are shown in bold. Total expresses marginal potential outcome probabilities.}
    \centering 
        \scalebox{1.}{
        \begin{tabular}{c|ccc|c}
            \toprule 
            \multicolumn{5}{c}{$\pr(\Yz, \Yo \mid X = 0)$} \\ \midrule
        \diagbox{$\Yz$}{$\Yo$}& -1        & 0    & 1        & Total \\ \midrule 
                            -1& {\bf 0.5} & 0    & 0        & {\bf 0.5} \\
                             0&        0  & 0    & 0        & 0 \\
                             1&        0  & 0    & {\bf 0.5}& {\bf 0.5} \\ \midrule
                         Total& {\bf 0.5} & 0    & {\bf 0.5}& {\bf 1.0} \\
            \bottomrule
        \end{tabular}
        } 
        \scalebox{1.}{
        \begin{tabular}{c|ccc|c}
            \toprule 
            \multicolumn{5}{c}{$\pr(\Yz, \Yo \mid X = 1)$} \\ \midrule
        \diagbox{$\Yz$}{$\Yo$}& -1        & 0    & 1        & Total \\ \midrule 
                            -1&        0  & 0    & {\bf 0.5}& {\bf 0.5} \\
                             0&        0  & 0    & 0        & 0 \\
                             1& {\bf 0.5} & 0    & 0        & {\bf 0.5} \\ \midrule
                         Total& {\bf 0.5} & 0    & {\bf 0.5}& {\bf 1.0} \\
            \bottomrule
        \end{tabular}
        }
\label{UAI2022-table_ex_2}

\end{table}

\section{Characteristic kernels} \label{asec:kernel}

This section provides a brief overview on characteristic kernels. For the formal definition, see e.g., \citet{sriperumbudur2010hilbert} and \citet[Section 3.3.1]{muandet2017kernel}.

The notion of characteristic kernels is closely related to \textit{kernel mean embedding} \citep{smola2007hilbert}, which is defined as the mean of feature mapping induced by a kernel function. Let $k_X\colon \mathcal{X} \times \mathcal{X} \rightarrow \R$ be a symmetric and positive-definite kernel function and $\Phi_X(x) \coloneqq k_X(x, \cdot)$ be the feature mapping of kernel $k_X$ that maps point $x \in \mathcal{X}$ into reproducing kernel Hilbert space (RKHS) $\mathcal{H}_{k_X}$. Then kernel mean embedding is defined as the mean of random variable $\Phi_X(X)$:
\begin{align*}
    \mu_{X} \coloneqq \E_{X}[\Phi_X(X)] \in \mathcal{H}_{k_X}.
\end{align*}
Here, the expectation is taken with respect to distribution $\pr(X)$; therefore, the concept of kernel mean embedding can be regarded as a mapping of distribution $\pr(X)$ into the RKHS, i.e., $\pr(X) \mapsto \mu_X \in \mathcal{H}_{k_X}$.

A characteristic kernel is a kernel function whose kernel mean embedding does not map different distributions to the same point in the RKHS; that is, the mapping by kernel mean embedding is injective \citep{sriperumbudur2010hilbert}. 

Roughly speaking, a kernel function is characteristic if mean $\E_{X}[\Phi_X(X)]$ contains all moments of random variable $X$. For instance, Gaussian kernel $k_X(x, x') = \mathrm{exp}(- \frac{ (x - x')^2}{2 h_X^2})$ for $x, x' \in \mathbb{R}^1$ is characteristic because the feature mapping is given as $\Phi_X(x) = \mathrm{e}^{- x^2 / 2 h_X^2} [1, \sqrt{\frac{1}{1! h_X^2}}x, \sqrt{\frac{1}{2! h_X^4}}x^2, \dots ]^{\top}$, and its expected value $\mathbb{E}_X[\Phi_X(X)]$ includes all moments: $\mathbb{E}_X[X], \mathbb{E}_X[X^2], \dots$.

By contrast, if $k_X$ is given as a polynomial function (i.e., polynomial kernel), $k_X$ is \textbf{not} a characteristic kernel. For instance, if $k_X$ is formulated as the 2nd-order polynomial kernel $k_X(x, x') = (1 + xx')^2$ for $x, x' \in \mathbb{R}^1$, the feature mapping is given as the finite-dimensional vector $\Phi_X(x) = [1, \sqrt{2} x, x^2]$. In this case, no element in expectation $\mathbb{E}_X [\Phi_X(X)]$ is represented as a function of higher-order moments than $2$; hence, kernel $k_X$ is not characteristic.

\section{Proofs}

\subsection{Proposition 1} \label{asubsec:prop1}

\begin{proof}
    Recall the following definition of $\mbox{WCMMD}^2_{X_{\si} = \xn}$:
    \begin{align} 
        &\mbox{WCMMD}^2_{X_{\si} = \xn} \nonumber \\
        \coloneqq &\E_{A, A', \biX_{-\si}, \biX'_{-\si}, Y, Y' \mid X_{\si} = X'_{\si} = \xn}[ \wpz(A, \biX) \wpz(A', \biX') k_Y(Y, Y')] \nonumber \\
        + &\E_{A, A', \biX_{-\si}, \biX'_{-\si}, Y, Y' \mid X_{\si} = X'_{\si} = \xn}[\wpo(A, \biX) \wpo(A', \biX') k_Y(Y, Y')] \nonumber \\
        - &2 \E_{A, A', \biX_{-\si}, \biX'_{-\si}, Y, Y' \mid X_{\si} = X'_{\si} = \xn}[\wpz(A, \biX) \wpo(A', \biX') k_Y(Y, Y')]. \tag{\ref{UAI2022-eq-Dj}}
    \end{align}

    We show that the first term in \eqref{UAI2022-eq-Dj} equals the one in $D^2_{\si}(\xn)$ in \eqref{UAI2022-eq-CDTE}. Using conditional ignorability and positivity assumptions, we have
    \begin{align*}
        &\E_{A, A', \biX_{-\si}, \biX'_{-\si}, Y, Y' \mid X_{\si} = \xn, X'_{\si} = \xn}[ \wpz(A, \biX) \wpz(A', \biX') k_Y(Y, Y')] \\
        = &\E_{\biX_{-\si}, \biX'_{-\si} \mid X_{\si} = \xn, X'_{\si} = \xn}\left[ \E_{A, A', Y, Y' \mid \biX_{-\si}, \biX'_{-\si}, X_{\si} = \xn, X'_{\si} = \xn}\left[\frac{\I(A=0)}{1 - \mathrm{e}(\biX)} \frac{\I(A'=0)}{1 - \mathrm{e}(\biX')} k_Y(Y, Y') \right] \right]\\
        = &\E_{\biX_{-\si}, \biX'_{-\si} \mid X_{\si} = \xn, X'_{\si} = \xn, A = 0, A' =0}\left[ \E_{\Yz, \yzp \mid \biX_{-\si}, \biX'_{-\si}, X_{\si} = \xn, X'_{\si} = \xn, A = 0, A' = 0}\left[ \frac{\pr(A=0)}{\pr(A=0 \mid \biX)} \frac{\pr(A'=0)}{\pr(A'=0 \mid \biX'))}  k_Y(Y, Y') \right] \right] \\
        = &\E_{\biX_{-\si}, \biX'_{-\si} \mid X_{\si} = \xn, X'_{\si} = \xn}[ \E_{\Yz, \yzp \mid \biX_{-\si}, \biX'_{-\si}, X_{\si} = \xn, X'_{\si} = \xn}[k_Y(\Yz, \yzp)] ] \\ 
        = &\E_{\Yz, \yzp \mid X_{\si} = \xn, X'_{\si} = \xn}[k_Y(\Yz, \yzp)].
    \end{align*}
    Similarly, the second and third terms in Eq. \eqref{UAI2022-eq-Dj} equal those in $\mathrm{MMD}^2(\pr(\Yz \mid \xn), \pr(\Yo \mid \xn))$ in Eq. \eqref{UAI2022-eq-CDTE}. Thus we proved \Cref{UAI2022-thm_prop1}.
\end{proof}

\subsection{Theorem 1} \label{asubsec:thm1}

From \Cref{UAI2022-thm_prop1}, we only have to show that $\widehat{D}^2_{\si}(\xn) \overset{p}{\rightarrow} \mbox{WCMMD}^2_{X_{\si} = \xn}$ ($n \rightarrow \infty$) under the assumptions of conditional ignorability and positivity:
\begin{assumption}[Conditional ignorability] \label{UAI2022-asmp1}
    For treatment $A$, features $\biX$, and potential outcomes $\Yz$ and $\Yo$, the following conditional independence relation holds:
    \begin{align*}
        \{\Yz, \Yo\} \indep A \mid \biX.
    \end{align*} 
\end{assumption}
\begin{assumption}[Positivity] \label{UAI2022-asmp2}
    For any value $\bix$ of features $\biX$, propensity score $\mathrm{e}(\biX)$ satisfies the following support condition:
    \begin{align*}
        0 < \mathrm{e}(\bix) < 1.
    \end{align*}
\end{assumption}

To prove $\widehat{D}^2_{\si}(\xn) \overset{p}{\rightarrow} \mbox{WCMMD}^2_{X_{\si} = \xn}$ ($n \rightarrow \infty$), we make several additional assumptions and impose the condition that the following symmetric function is square integrable: 
\begin{align*}
    &K((A, \biX, Y), (A', \biX', Y')) \nonumber \\
    \coloneqq &\left( \wpz(A, \biX) \wpz(A', \biX') + \wpo(A, \biX, Y) \wpo(A', \biX', Y') - \wpz(A, \biX) \wpo(A', \biX') - \wpo(A, \biX) \wpz(A', \biX') \right) k_Y(Y, Y'). 
\end{align*}
\begin{assumption} \label{UAI2022-asmp3}
    Symmetric function $K$ is square integrable:
    \begin{align*}
        \E_{A, A', \biX, \biX', Y, Y'}[K((A, \biX, Y), (A', \biX', Y'))] < \infty.
    \end{align*}
\end{assumption}
When $X_{\si}$ is continuous-valued, and $\oax$ is given by \eqref{UAI2022-eq-oax_cont}, we make the following standard assumptions on kernel function $k_{X_{\si}}$:
\begin{assumption} \label{UAI2022-asmp4}
    Let $K_{X_{\si}}$ be the following kernel function that measures the similarity between two values $x_{\si}$ and $x^{\star}_{\si}$ on $\mathcal{X}$:
    \begin{align*}
     K_{X_{\si}}(x_{\si} - x^{\star}_{\si}) \coloneqq \frac{1}{h_{X_{\si}}} k_{X_{\si}}(x_{\si}, x^{\star}_{\si}).
    \end{align*}
    Then the order of function $K_{X_{\si}}(u)$ is given by integer $\delta \geq 2$; in other words, the following holds:
    \begin{align*}
        \int u^{\delta} K_{X_{\si}}(u) du < \infty.
    \end{align*}
\end{assumption}
\begin{assumption} \label{UAI2022-asmp5}
    Bandwidth $h_{X_{\si}}$ of kernel function $k_{X_{\si}}$ satisfies 
    \begin{align*}
       h_{X_{\si}} \rightarrow 0 \quad \mbox{and} \quad nh_{X_{\si}} \rightarrow \infty. \quad (n \rightarrow \infty)
    \end{align*}
\end{assumption}
In addition, we impose the smoothness conditions on marginal distribution $\pr(X_{\si})$ and the joint distribution of features $\pr(\biX)$:
\begin{assumption} \label{UAI2022-asmp6}
    Density functions $\pr(X_{\si})$ and $\pr(\biX)$ are $\delta$ times continuously differentiable.
\end{assumption}

Using these assumptions, we prove \Cref{UAI2022-thm_thm1}:

\begin{proof}
    \textbf{The case where weight $\oax_i$ is given by Eq. \eqref{UAI2022-eq-oax_disc}:} Let $K_{i, j} \coloneqq K((a_i, \bix_i, y_i), (a_j, \bix_j, y_j))$ for $i, j \in \{1, \dots, n\}$ and $n_{\xn} \coloneqq \sum_{i=1}^n \I(x_{\si, i} = \xn)$. Then empirical estimator $\widehat{D}^2_{\si}(\xn)$ is given as
    \begin{align*}
        \widehat{D}^2_{\si}(\xn) &= \frac{1}{n^2_{\xn}} \sum_{i=1}^n \sum_{j=1}^n \I(x_{\si, i} = \xn) \I(x_{\si, j} = \xn) K_{i, j}  \\
        &= \left( \frac{n}{n_{\xn}} \right)^2  \frac{1}{n^2} \sum_{i=1}^n \sum_{j=1}^n \I(x_{\si, i} = \xn) \I(x_{\si, j} = \xn) K_{i, j}  \\
        &=  \left( \frac{n}{n_{\xn}} \right)^2 V_n^{\xn},
    \end{align*}
    where 
    \begin{align*}
        V_n^{\xn} \coloneqq \frac{1}{n^2} \sum_{i=1}^n \sum_{j=1}^n \I(x_{\si, i} = \xn) \I(x_{\si, j} = \xn) K_{i, j}
    \end{align*}
    is a V-statistic whose corresponding U-statistic is given as
    \begin{align*}
        U_n^{\xn} \coloneqq \frac{1}{{}_n \mathrm{C}_2} \sum_{i < j} \I(x_{\si, i} = \xn) \I(x_{\si, j} = \xn) K_{i, j}.
    \end{align*}

    We prove the consistency of $\widehat{D}^2_{\si}(\xn)$ by showing the following three relations: 
    \begin{align}
        &U_n^{\xn} \overset{a.s.}{\rightarrow} \E_{A, A', \biX, \biX', Y, Y'}[\I(X_{\si} = \xn) \I(X_{\si} = \xn)  K((A, \biX, Y), (A', \biX', Y'))] \label{UAI2022-proof_disc1} \\
        &\left( \frac{n}{n_{\xn}} \right)^2 U_n^{\xn} \overset{a.s.}{\rightarrow} \mbox{WCMMD}^2_{X_{\si} = \xn} \label{UAI2022-proof_disc2} \\
        &U_n^{\xn} - V_n^{\xn} \overset{p}{\rightarrow} 0 \label{UAI2022-proof_disc3}. 
    \end{align}
    Relation \eqref{UAI2022-proof_disc1} holds from the Strong Law of Large Numbers for U-statistics \citep{hoeffding1961strong}. By combining this relation with the fact that $\frac{n_{\xn}}{n} = \frac{1}{n} \sum_{i=1}^n \I(x_{\si, i} = \xn)  \overset{a.s.}{\rightarrow} \pr(X_{\si} = \xn)$, we can derive the relation in Eq. \eqref{UAI2022-proof_disc2}. The relation in Eq. \eqref{UAI2022-proof_disc3} can be shown as follows. Under \Cref{UAI2022-asmp3}, since $\E[K((A, \biX, Y), (A', \biX', Y'))] \leq \E[K((A, \biX, Y), (A, \biX, Y))] < \infty$, by employing Lemma 5.7.3 in \citet{serfling2009approximation}, we have $\E[| U_n^{\xn} - V_n^{\xn} |] = O(n^{-1})$,  and thus by applying Markov's inequality, we have 
    \begin{align*}
        \pr(| U_n^{\xn} - V_n^{\xn} | \geq \epsilon) \leq \frac{\E[| U_n^{\xn} - V_n^{\xn} |]}{\epsilon} \rightarrow 0 \quad \mbox{as $n \rightarrow \infty$},
    \end{align*}
    which is sufficient to prove the relation in Eq. \eqref{UAI2022-proof_disc3}.

    By combining Eq. \eqref{UAI2022-proof_disc1}, \eqref{UAI2022-proof_disc2}, and \eqref{UAI2022-proof_disc3}, we have $\widehat{D}^2_{\si}(\xn) \overset{p}{\rightarrow} \mbox{WCMMD}^2_{X_{\si} = \xn}$ as $n \rightarrow \infty$. Since \Cref{UAI2022-thm_prop1} holds under \Cref{UAI2022-asmp1,UAI2022-asmp2}, we have $\widehat{D}^2_{\si}(\xn) \overset{p}{\rightarrow} D^2_{\si}(\xn)$ as $n \rightarrow \infty$. Thus we prove the consistency of $\widehat{D}^2_{\si}(\xn)$.

    \textbf{The case where weight $\oax_i$ is given by Eq. \eqref{UAI2022-eq-oax_cont}:}

    In this case, empirical estimator $\widehat{D}^2_{\si}(\xn)$ is given as
    \begin{align}
        \widehat{D}^2_{\si}(\xn) &= \frac{\frac{1}{n^2 h^2_{X_{\si}}} \sum_{i=1}^n \sum_{j=1}^n k_{X_{\si}} (x_{\si, i}, \xn) k_{X_{\si}} (x_{\si, j}, \xn) K_{i, j}}{\frac{1}{n^2 h^2_{X_{\si}}} \sum_{i=1}^n \sum_{j=1}^n k_{X_{\si}} (x_{\si, i} , \xn) k_{X_{\si}} (x_{\si, j} , \xn)}. \label{UAI2022-proof_cont}
    \end{align}

    From the Strong Law of Large Numbers, as $n \rightarrow \infty$, the numerator in Eq. \eqref{UAI2022-proof_cont} converges to the following expected value:
    \begin{align*}
        \E_{A, A', \biX, \biX', Y, Y'}\left[ \frac{1}{h^2_{X_{\si}}} K_{X_{\si}} \left(\frac{X_{\si} - \xn}{h_{X_{\si}}}\right) K_{X_{\si}} \left(\frac{X'_{\si} - \xn}{h_{X_{\si}}}\right) K((A, \biX, Y), (A', \biX', Y'))\right].
    \end{align*}
    Under \Cref{UAI2022-asmp4,UAI2022-asmp6}, we can reformulate this expected value by performing a Taylor expansion as follows:
    \begin{align}
        &\E_{A, A', \biX, \biX', Y, Y'}\left[ \frac{1}{h^2_{X_{\si}}} K_{X_{\si}} \left(\frac{X_{\si} - \xn}{h_{X_{\si}}}\right) K_{X_{\si}} \left(\frac{X'_{\si} - \xn}{h_{X_{\si}}}\right) K((A, \biX, Y), (A', \biX', Y'))\right] \nonumber \\
        = &\E_{U=u, V=v}[ \E_{A, A', \biX_{-\si}, \biX'_{-\si}, Y, Y' \mid X_m = \xn + h_{X_{\si}} u, X'_m = \xn + h_{X_{\si}} v}[ \pr(X_{\si} = \xn + h_{X_{\si}} u) \pr(X'_{\si} = \xn + h_{X_{\si}} v) K_{X_{\si}}(u) K_{X_{\si}}(v) K((A, \biX, Y), (A', \biX', Y')) ]] \nonumber \\
        = &\E_{A, A', \biX_{-\si}, \biX'_{-\si}, Y, Y' \mid X_m = \xn, X'_m = \xn}[\pr^2(X_{\si} = \xn) K((A, \biX, Y), (A', \biX', Y'))] + O_p \left(h_{X_{\si}}^{\delta}\right). \label{UAI2022-proof_cont1}
    \end{align}

    Regarding the denominator in Eq. \eqref{UAI2022-proof_cont}, from the consistency results of the kernel density estimator in \citet{wied2012consistency}, we have
    \begin{align}
        \frac{1}{n h_{X_{\si}}} \sum_{j=1}^n k_{X_{\si}} (x_{m, j}, \xn) \overset{a.s.}{\rightarrow} \pr(X_{\si} = \xn). \label{UAI2022-proof_cont2}
    \end{align}

    By combining Eqs. \eqref{UAI2022-proof_cont1} and \eqref{UAI2022-proof_cont2}, under \Cref{UAI2022-asmp5}, we have $\widehat{D}^2_{\si}(\xn) \overset{p}{\rightarrow} \mbox{WCMMD}^2_{X_{\si} = \xn}$ as $n \rightarrow \infty$. Using \Cref{UAI2022-thm_prop1}, we have $\widehat{D}^2_{\si}(\xn) \overset{p}{\rightarrow} D^2_{\si}(\xn)$ as $n \rightarrow \infty$. Thus we proved the consistency of $\widehat{D}^2_{\si}(\xn)$. 

\end{proof}

\section{Additional Experimental Results}

In what follows, we present several additional synthetic data experiments to further evaluate the performance of our method. \Cref{asubsec:counterexp} shows the performance on the data where the truly relevant features do not affect the discrepancy between marginal potential outcome distributions, which is our inference target. \Cref{asubsec:neuronexp} displays the results when using different neural network architectures in the models of propensity score and CVAE.

\subsection{Examining Counterexamples} \label{asubsec:counterexp}

This section presents the performance of our method on the synthetic data where the features do not influence the discrepancy between conditional distributions $\pr(\Yz \mid X_{\si})$ and $\pr(\Yo \mid X_{\si})$ but affect joint distribution $\pr(\Yz, \Yo \mid X_{\si})$. With such data, our method does not work well because it relies on the discrepancy between $\pr(\Yz \mid X_{\si})$ and $\pr(\Yo \mid X_{\si})$, as described in \Cref{UAI2022-sec3-1}.

To evaluate the performance, we prepared synthetic data in a similar manner to \Cref{UAI2022-sec4-2}, which only differs in the generation process of potential outcomes $\Yz$ and $\Yo$. Here, we set the sample size to $n=2000$ and sampled the values of $\Yz$ and $\Yo$ from the following $2$-dimensional Gaussian distributions:
\begin{itemize}
    \item \textbf{LinCovar}:
\begin{align}
    \left[
        \begin{array}{c}
            \Yz \\
            \Yo
        \end{array} \right]
    \sim \mathcal{N}\left(
        \left[
            \begin{array}{c}
                -5 \\
                0
            \end{array} \right],
            \left[
                \begin{array}{cc}
                    1 & 1 - \frac{1}{h(f(X_1,\dots,X_5))}\\
                    1 - \frac{1}{h(f(X_1,\dots,X_5))} & 1
                \end{array} \right]                    
       \right),     
\end{align}
    \item \textbf{NonlinCovar}:
    \begin{align}
        \left[
            \begin{array}{c}
                \Yz \\
                \Yo
            \end{array} \right]
        \sim \mathcal{N}\left(
            \left[
                \begin{array}{c}
                    -5 \\
                    0
                \end{array} \right],
                \left[
                    \begin{array}{cc}
                        1 & 1 - \frac{1}{h(g(X_1,\dots,X_5))}\\
                        1 - \frac{1}{h(g(X_1,\dots,X_5))} & 1
                    \end{array} \right]                    
           \right), 
    \end{align}    
\end{itemize}
where functions $f$, $g$, and $h$ are presented in \Cref{UAI2022-sec4-2}. Under LinCovar and NonlinCovar, features $X_1, \dots, X_5$ only influence the covariance between potential outcomes $\Yz$ and $\Yo$ and do not affect any functionals of the marginal distributions.

We performed $50$ experiments and evaluated their mean and standard deviation of TPRs and FPRs. \Cref{UAI2022-table_ex_counter} presents the results. As expected, our method could not correctly select features $X_1, \dots, X_5$ because their values do not affect the discrepancy between conditional potential outcome distributions.

\begin{table}[t]
    \caption{TPRs and FPRs of our method on LinCovar and NonlinCovar datasets. Mean and standard deviation over $50$ runs are shown.}
    \centering 
        \scalebox{1.}{
        \begin{tabular}{c|cc}
            \toprule 
                              & TPR        & FPR \\ \midrule 
                            LinCovar & 0.02 $\pm$ 0.06 & 0.02 $\pm$ 0.02 \\
                            NonlinCovar & 0.04 $\pm$ 0.08  & 0.02 $\pm$ 0.02  \\
            \bottomrule
        \end{tabular}
        } 

\label{UAI2022-table_ex_counter}

\end{table}

Note, however, that selecting these features is extremely challenging because it is impossible to estimate the covariance since we cannot infer the joint distribution of potential outcomes, as described in \Cref{UAI2022-sec3-1}. Due to this difficulty, all of the existing mean-based methods also fail, and compared with such methods, ours can detect a wider variety of features.

\subsection{Performance Evaluation with Different Neural Network Architectures} \label{asubsec:neuronexp}

Since our method relies on two neural network models to represent propensity function $\mathrm{e}(\biX)$ and CVAE $\mathcal{L}(X_{\si} \mid \biX_{-\si})$ ($\si = 1, \dots, \nf$), we confirmed how greatly the neural network architectures affect the overall feature selection performance.

For this purpose, we performed additional synthetic data experiments with sample size $n = 1000$. We evaluated the mean and standard deviation of TPRs and FPRs over $50$ runs by changing the number of neurons of each layer in the two-layered neural network models, which is fixed to $50$ for propensity score and to $128$ for CVAE in the experiments in \Cref{UAI2022-sec4-2}.

\Cref{UAI2022-table_neurons_propensity,UAI2022-table_neurons_cvae} display the results. With all synthetic datasets, the number of neurons in propensity score and CVAE did not greatly affect the performance. 

\newpage

\begin{table}[t]
    \caption{TPRs and FPRs of our method with different numbers of neurons in propensity score model. Mean and standard deviation over $50$ runs are shown.}
    \centering 
    \begin{tabular}{llcccc}
        \toprule
        &  & \multicolumn{4}{c}{Number of neurons in propensity score model}           \\

                   & & 25            & 50            & 100           & 200           \\ \midrule
    \multirow{2}{*}{LinMean}    & TPR     & 0.80$\pm$0.21 & 0.79$\pm$0.22 & 0.84$\pm$0.14 & 0.84$\pm$0.16 \\
                                & FPR     & 0.06$\pm$0.06 & 0.06$\pm$0.07 & 0.08$\pm$0.06 & 0.08$\pm$0.06 \\ 
    \multirow{2}{*}{NonlinMean} & TPR     & 0.95$\pm$0.10 & 0.94$\pm$0.12 & 0.98$\pm$0.06 & 0.97$\pm$0.08             \\
                                & FPR     & 0.04$\pm$0.04 & 0.04$\pm$0.04 & 0.03$\pm$0.03 & 0.05$\pm$0.04  \\
    \multirow{2}{*}{LinVar}     & TPR     & 0.71$\pm$0.19 & 0.73$\pm$0.19 & 0.77$\pm$0.16 & 0.76$\pm$0.18    \\
                                & FPR     & 0.08$\pm$0.07 & 0.07$\pm$0.08 & 0.10$\pm$0.07 & 0.09$\pm$0.07    \\
    \multirow{2}{*}{NonlinVar}  & TPR     & 0.64$\pm$0.25 & 0.62$\pm$0.25 & 0.63$\pm$0.26 & 0.64$\pm$0.25    \\
                                & FPR     & 0.04$\pm$0.04 & 0.04$\pm$0.04 & 0.04$\pm$0.04 & 0.04$\pm$0.04  \\ \bottomrule
    \end{tabular}
    \label{UAI2022-table_neurons_propensity}
\end{table}

\begin{table}[t]
    \caption{TPRs and FPRs of our method with different numbers of neurons in CVAE model. Mean and standard deviation over $50$ runs are shown.}
    \centering     
    \begin{tabular}{llcccc}
        \toprule
        &  & \multicolumn{4}{c}{Number of neurons in CVAE model}           \\
        &  & 16            & 64            & 128           & 256           \\ \midrule
        \multirow{2}{*}{LinMean}    & TPR     & 0.82$\pm$0.18 & 0.82$\pm$0.17 & 0.79$\pm$0.22 & 0.83$\pm$0.16 \\
                                & FPR     & 0.08$\pm$0.06 & 0.07$\pm$0.06 & 0.06$\pm$0.07 & 0.10$\pm$0.07 \\ 
    \multirow{2}{*}{NonlinMean} & TPR     & 0.96$\pm$0.09 & 0.98$\pm$0.06 & 0.94$\pm$0.12 & 0.94$\pm$0.05 \\
                                & FPR     & 0.04$\pm$0.04 & 0.03$\pm$0.03 & 0.04$\pm$0.04 & 0.05$\pm$0.04  \\
    \multirow{2}{*}{LinVar}     & TPR     & 0.68$\pm$0.19 & 0.66$\pm$0.17 & 0.73$\pm$0.19 & 0.70$\pm$0.16    \\
                                & FPR     & 0.07$\pm$0.05 & 0.06$\pm$0.05 & 0.07$\pm$0.08 & 0.08$\pm$0.07   \\
    \multirow{2}{*}{NonlinVar}  & TPR     & 0.58$\pm$0.25 & 0.56$\pm$0.25 & 0.62$\pm$0.25 & 0.60$\pm$0.20  \\
                                & FPR     & 0.02$\pm$0.03 & 0.03$\pm$0.03 & 0.04$\pm$0.04 & 0.04$\pm$0.05  \\ \bottomrule
    \end{tabular}
    \label{UAI2022-table_neurons_cvae}
\end{table}

\end{document}